\documentclass[runningheads]{llncs}
\usepackage{graphicx}
\usepackage{amsmath,amssymb} 
\usepackage{color}

\usepackage{booktabs}
\usepackage{tabularx}
\usepackage[export]{adjustbox}

\newcommand{\myparagraph}[1]{\textbf{\textit{#1}}}

\begin{document}
\title{Explainable Neural Computation via Stack Neural Module Networks}

\titlerunning{Explainable Neural Computation via Stack Neural Module Networks}
%
\author{Ronghang Hu\inst{1} \and Jacob Andreas\inst{1} \and Trevor Darrell\inst{1} \and Kate Saenko\inst{2}}
%
\authorrunning{R. Hu, J. Andreas, T. Darrell, K. Saenko}
%

\institute{$^1$University of California, Berkeley \qquad $^2$Boston University \\
\email{\{ronghang,jda,trevor\}@eecs.berkeley.edu, saenko@bu.edu}
}
\maketitle              
\begin{abstract}
In complex inferential tasks like question answering, machine learning models must confront two challenges: the need to implement a compositional \emph{reasoning} process, and, in many applications, the need for this reasoning process to be \emph{interpretable} to assist users in both development and prediction. Existing models designed to produce interpretable traces of their decision-making process typically require these traces to be supervised at training time. In this paper, we present a novel neural modular approach that performs compositional reasoning by automatically inducing a desired sub-task decomposition without relying on strong supervision. Our model allows linking different reasoning tasks though shared modules that handle common routines across tasks. Experiments show that the model is more interpretable to human evaluators compared to other state-of-the-art models: users can better understand the model's underlying reasoning procedure and predict when it will succeed or fail based on observing its intermediate outputs.
\end{abstract}

\section{Introduction}

Deep neural networks have achieved impressive results on many vision and language tasks. Yet the predictive power of generic deep architectures comes at a cost of lost interpretability, as these architectures are essentially black boxes with respect to human understanding of their predictions. This can impair human users' trust in learning systems and make them harder to refine \cite{doshi2017towards}.

\begin{figure}[t]
\center
\includegraphics[width=\linewidth]{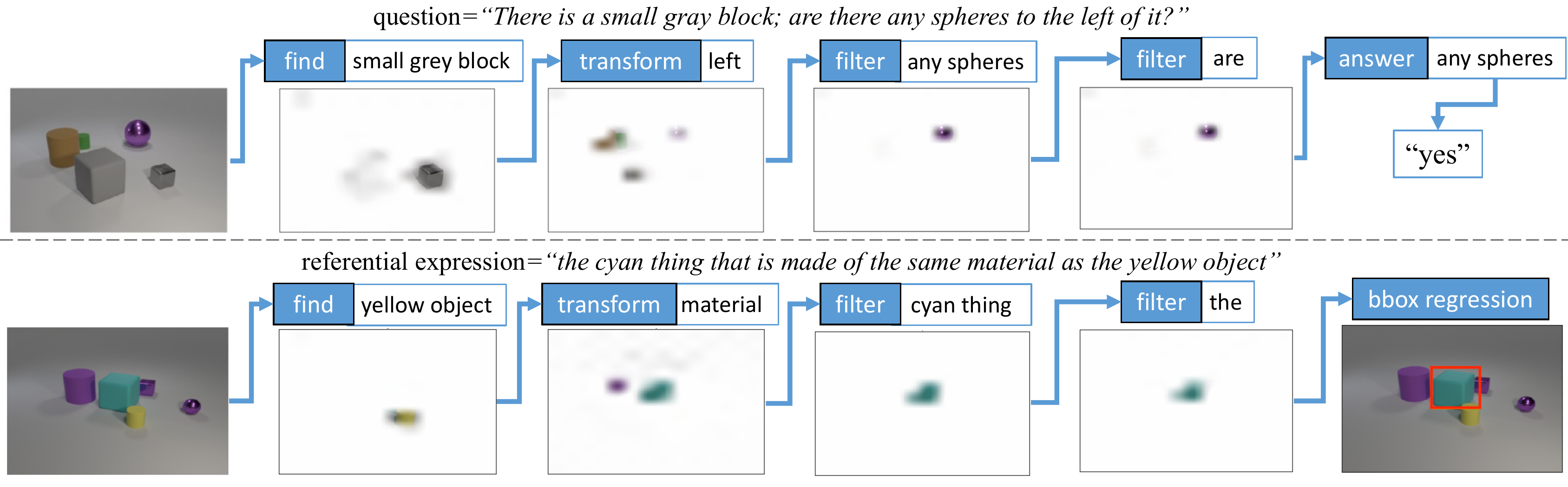}
\caption{Our model reveals interpretable subtask structure by inducing a decomposition of the reasoning procedure into several sub-tasks, each addressed by a neural module. It can simultaneously answer visual questions and ground referential expressions.}
\label{fig:teaser}
\end{figure}

These issues have led to recent efforts in explaining neural models, ranging from building in attention layers to post-hoc extraction of implicit model attention, e.g.\ by gradient propagation \cite{springenberg2014striving,zhang2016top,ramanishka2017top,zhou2016learning,ramprasaath2016grad}, post-hoc natural language explanations \cite{hendricks2016generating,andreas2017translating} and network dissection \cite{bau2017network}. Such approaches can highlight the image regions that are most important for predicting a particular label or provide a textual interpretation of the network output. However, explainable models of more complex problems involving multiple sub-tasks, such as Visual Question Answering (VQA) \cite{antol2015vqa} and Referential Expression Grounding (REF) \cite{rohrbach2016grounding}, are less studied in comparison.
Complex problems may require several reasoning steps to solve. For example in Figure~\ref{fig:teaser}, the question ``There is a small gray block; are there any spheres to the left of it?" might require solving the following subtasks: find the ``small gray block", look for ``spheres to the left of it" and decide whether such object exists in the image. Therefore, a single heat-map highlighting important spatial regions such as \cite{ramprasaath2016grad} may not tell the full story of how a model performs.

In this paper, we present a new model that makes use of an explicit, modular reasoning process, but which allows fully differentiable training with back-propagation and without expert supervision of reasoning steps. 
Existing modular networks first analyze the question and then predict a sequence of pre-defined modules (each implemented as a neural net) that chain together to predict the answer. However, they need an ``expert layout'', or supervised module layouts for training the layout policy in order to obtain good accuracy. 
Our proposed approach, the \textit{Stack Neural Module Network} or \textit{Stack-NMN}, can be trained without layout supervision, and replaces the layout graph of \cite{hu2017learning} with a stack-based data structure. Instead of making discrete choices on module layout, in this work we make the layout soft and continuous, so that our model can be optimized in a fully differentiable way using gradient descent. We show that this improves both the accuracy and interpretability compared to existing modular approaches.
We also show that this model can be extended to handle both Visual Question Answering (VQA) \cite{antol2015vqa} and Referential Expression Grounding (REF) \cite{rohrbach2016grounding} seamlessly in a single model by sharing knowledge across related tasks through common routines as in Figure~\ref{fig:teaser}.

A variety of different model architectures have been proposed for complex reasoning and question answering.
Our evaluation in this paper focuses on both the accuracy and interpretability of these models. In particular, we ask:
\textit{does explicit modular structure make models more interpretable?}
We use the CLEVR dataset \cite{johnson2017clevr} as a testbed, as it poses a task of high complexity.
State-of-the-art models for this task vary in the degree to which they provide ``explanations''. Relation Networks \cite{santoro2017simple} and FiLM \cite{perez2018film} achieve high performance but do not expose their internal decision process. Other state-of-the-art models on CLEVR use recurrent layers to compute the answer over multiple steps and output different image and/or text attention at each step. These include modular networks \cite{andreas16neural,andreas2016learning,hu2017learning,johnson2017inferring,mascharka2018transparency}, and non-modular recurrent attention models \cite{yang2016stacked,hudson2018compositional}. It has been suggested by the authors that the attention and/or module layouts inferred by these methods can be regarded as explanations of the networks' internal reasoning process. Yet, to the best of our knowledge, their meaningfulness to humans has never been explicitly evaluated;
we provide a more rigorous assessment of the interpretability of multi-step attentional VQA models here.

We categorize existing multi-step models in terms of whether they have a discrete library of structured modules for each step (e.g., NMN and related approaches \cite{andreas16neural,andreas2016learning,hu2017learning,johnson2017inferring,andreas2017modular,mascharka2018transparency}), vs.\ homogeneous subtask computational elements (e.g., multi-hop attention networks \cite{yang2016stacked,xu2016ask}, MAC \cite{hudson2018compositional}, etc.).
We assess these models below and identify tradeoffs between accuracy and interpretability of these existing model classes. We find that our proposed Stack-NMN model has comparable performance to existing modular approaches even without expert supervision, while achieving the greatest interpretability among evaluated models with respect to both subjective and objective measures of human understanding.

\section{Related work}

\myparagraph{Visual question answering (VQA).} The task of visual question answering is to infer the answer based on the input question and image. Existing methods on VQA can be mainly categorized into holistic approaches (e.g., \cite{yang2016stacked,xu2016ask,fukui2016multimodal,anderson2017bottom,santoro2017simple,perez2018film,hudson2018compositional}), and neural module approaches \cite{andreas16neural,andreas2016learning,hu2017learning,johnson2017inferring,mascharka2018transparency}. The major difference between these two lines of work is that neural module approaches explicitly decompose the reasoning procedure into a sequence of sub-tasks, and have specialized modules to handle the sub-tasks, while holistic approaches do not have explicit sub-task structure, and different kinds of reasoning routines are all handled homogeneously.

Some holistic models perform sequential interactions between the image and the question. For example, SAN \cite{yang2016stacked} uses multi-hop attention to extract information from the image. FiLM \cite{perez2018film} uses multiple conditional batch normalization layers to fuse the image representation and question representation. Among these methods, MAC \cite{hudson2018compositional} performs multiple steps of reading and writing operations to extract information from the image and update its memory. Although these models have sequential interactions between the input image and the question, they do not explicitly decompose the reasoning procedure into semantically-typed sub-tasks. In our model, we adopt a similar textual attention mechanism as in \cite{hudson2018compositional} in Sec.~\ref{sec:controller}, while also predicting module weights from the input text.

\myparagraph{Neural module networks (NMNs).} In NMN \cite{andreas16neural}, N2NMN \cite{hu2017learning}, PG+EE \cite{johnson2017inferring} and TbD \cite{mascharka2018transparency}, the inference procedure is performed by analyzing the question and decomposing the reasoning procedure into a sequence of sub-tasks. In \cite{hu2017learning}, \cite{johnson2017inferring} and \cite{mascharka2018transparency}, a layout policy is used to turn the question into a module layout. Then the module layout is executed with a neural module network. Here, given an input question, the layout policy learns what sub-tasks to perform, and the neural modules learn how to perform each individual sub-tasks.

However, it is shown in these previous work that ``expert layouts'' (i.e.\ human annotations of the desired layout) are needed to pretrain or supervise the layout policy in order to get compositional behavior and good accuracy. Without expert guidance, existing models suffer from significant performance drops or fail to converge. This indicates that it is challenging to simultaneously learn ``what'' and ``how'' in these models. In this work, we address this problem with soft and continuous module layout, making our model fully differentiable and trainable with using gradient descent without resorting to expert layouts.

\myparagraph{Interpretable reasoning and explainable neural networks.} Recent years have seen increased interest in various aspects of interpretability in learned models \cite{otte2013safe}, particularly in neural networks \cite{olah2018building}.
This includes work aimed at both explaining the decision rules implemented
by learned models, and the mechanisms by which these rules are derived from data \cite{selbst2018intuitive,koh2017understanding}. In the present work we are primarily interested
in the former. 
One line of research in this direction attempts to generate post-hoc explanations of decisions 
from generic model architectures, either by finding interpretable local surrogates in the form of 
linear models \cite{ribeiro2016should}, logical rules 
\cite{duch1998extraction,zhang2018interpreting} or natural language descriptions \cite{andreas2017translating,zhou2017interpreting}, or by visualizing salient features \cite{ramanishka2017top,ramprasaath2016grad}.

An alternative line of work investigates the extent to which models can be 
explicitly designed from the outset to provide enhanced interpretability, where main focus of study has been visual attention \cite{mnih2014recurrent,park2016attentive}. While
the various modular approaches described above
are sometimes described as ``interpretable'' \cite{hu2017learning}, we are not aware of any
research evaluating this in practice. In the present
work, our goal is to evaluate whether this kind of
explicit modular structure, and not just iterated
attention, improves interpretability in concrete evaluation scenarios.

\myparagraph{Multi-task learning.}
Different from existing multi-task approaches such as sharing common features (e.g., \cite{he2017mask}), our model simultaneously handles both Visual Question Answering (VQA) \cite{antol2015vqa} and Referential Expression Grounding (REF) \cite{rohrbach2016grounding} by exploiting the intuition that related tasks should have common sub-routines, and addressing them with a common set of neural modules.

\section{Approach}

\begin{figure}[t]
\centering
\includegraphics[trim={0 53mm 0 0},clip,width=\linewidth]{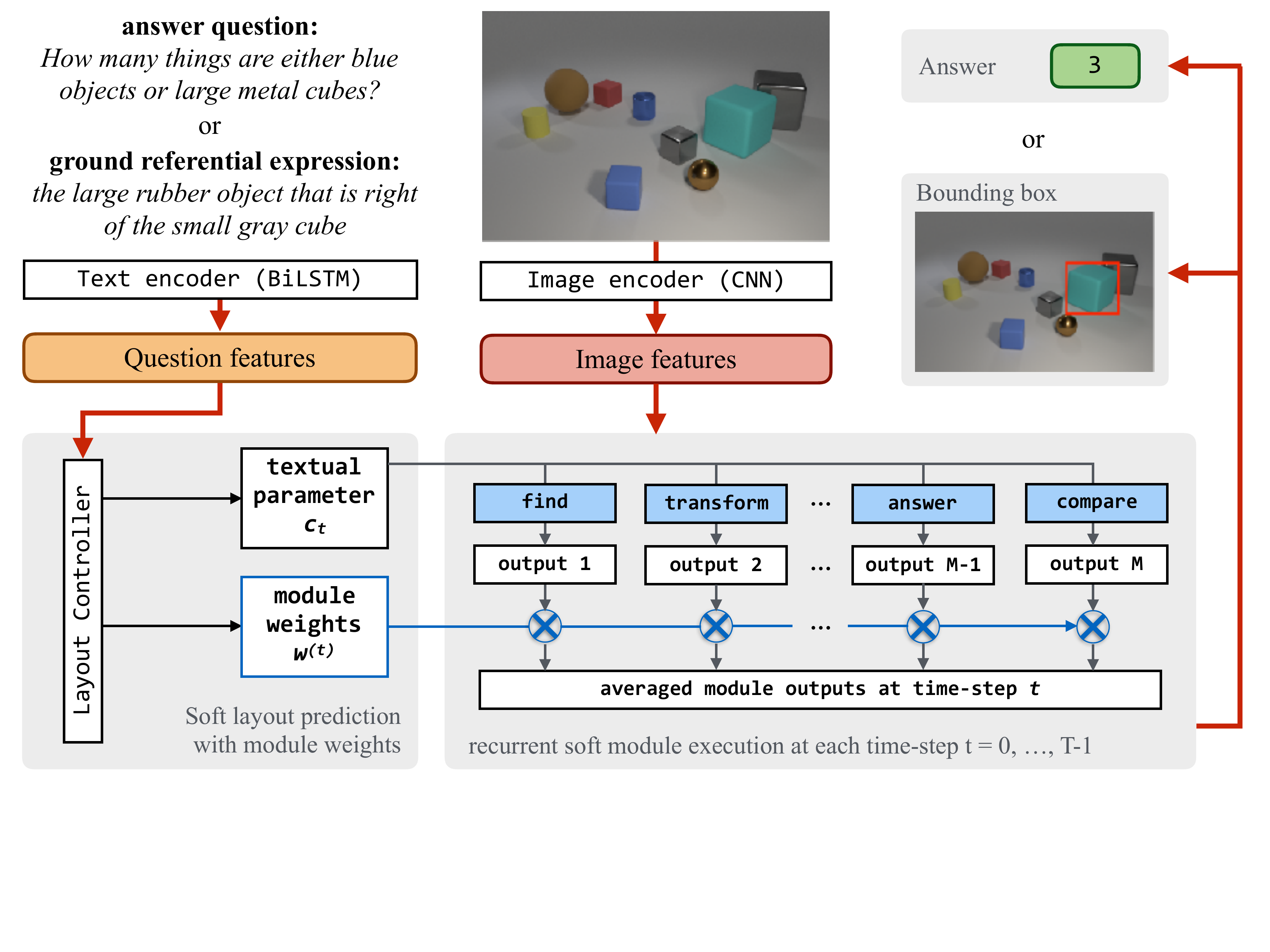}
\caption{Overview of our model. Our model predicts a continuous layout via module weights $w^{(t)}$ and executes the modules in a soft manner with a memory stack.}
\label{fig:method}
\end{figure}

In this paper, we analyze and design interpretable neural networks for high-complexity VQA and REF tasks. We evaluate the interpretability of multi-step VQA networks to humans, and in particular compare modular networks to non-modular networks in terms of how well humans can understand the internal computation process. We describe our proposed evaluation strategy and results in Section~\ref{sec:exp_compare}. We also improve modular networks by proposing a new formulation, which we describe in this section. Specifically, we describe Stack Neural Module Networks (Stack-NMNs) with the following components. 1) A layout controller that decomposes the reasoning task into a sequence of sub-tasks, and translates the input question into a \textit{soft layout}, specified via a soft distribution over module weights $w^{(t)}$ at each timestep $t$. The controller also supplies each module with a textual parameter $c_t$ at every time-step using textual attention. 2) A set of neural modules $M$ to handle the sub-tasks specified by the controller. Each neural module is a differentiable parameterized function that performs a specific sub-task, and can be executed dynamically on-the-fly according to the soft layout. 3) A differentiable memory stack to store and retrieve intermediate outputs from each module during execution.

Figure~\ref{fig:method} shows an overview of our model. The overall architecture of our model is conceptually similar to N2NMN \cite{hu2017learning}, where layout controller in our model resembles the previous layout policy.
The major difference between our model and this prior work lies in whether the layout selection is continuous or discrete. N2NMN makes discrete choices of module layout in a graph structure and can only be end-to-end optimized with reinforcement learning approaches. On the other hand, our model makes soft layout selection with a differentiable stack structure, by giving each module a continuous-valued weight parameter and averaging the outputs of all modules according to their weights. This makes the execution procedure fully differentiable so that our model is trainable with back-propagation like other neural networks.

\subsection{Module layout controller}
\label{sec:controller}

The layout controller in our model decides what subtask to perform at each execution time step $t$ by selecting a module $m_t$ for that time step, and also supplying it with a textual parameter $c_t$ to give specific instruction to the module $m_t \in M$. For example, the controller may decide to look for red things at $t=0$, by running a \texttt{Find} module with a textual parameter $c_t$ that contains the information for the word ``red''.

The structure of our layout controller is similar to the control unit in \cite{hudson2018compositional}. Suppose there are $S$ words in the input question. The layout controller first encodes the input question $q$ into a $d$-dimensional sequence $[h_1, \cdots, h_S]$ of length $S$ using a bi-directional LSTM as $[h_1, \cdots, h_S] = \text{BiLSTM}(q; \theta_\mathrm{BiLSTM})$, where each $h_s$ is the concatenation of the forward LSTM output and the backward LSTM output at the $s$-th input word.
Next, the controller runs in a recurrent manner from time-step $t = 0$ to time-step $t=T-1$. At each time-step $t$, it applies a time-step dependent linear transform to the question $q$, and linearly combines it with the previous $d$-dimensional textual parameter $c_{t-1}$ as $u = W_2 \left[W_1^{(t)} q + b_1; c_{t-1}\right] + b_2$, where $W_1^{(t)}$ and $W_2$ are $d\times d$ and $d \times 2d$ matrices respectively, and $b_1$ and $b_2$ are $d$-dimensional vectors. Unlike all other parameters in the layout controller, $W_1^{(t)}$ is not shared across different time steps.

To select the module to execute at the current time-step $t$, a small multi-layer perceptron (MLP) is applied to $u$ to predict a $|M|$-dimensional vector $w^{(t)}$ as $w^{(t)} = \mathrm{softmax}(\mathrm{MLP}(u; \theta_\mathrm{MLP}))$. The module weight $w^{(t)}$ contains the weight distribution over each module $m \in M$ and sums up to one (i.e. $\sum_{m \in M}^M w_m^{(t)} = 1$), which resembles a probability distribution or soft attention over the modules. It is used to select modules in each time-step $t$ in a continuous manner.

Finally, the controller predicts a textual parameter $c_t$ with a textual attention over the encoded question words as
$cv_{t,s} = \mathrm{softmax}(W_3 (u \odot h_s))$ and $c_t = \sum_{s=1}^S cv_{t,s} \cdot h_s$, where $\odot$ is element-wise multiplication, $W_3$ is a $1\times d$ matrix, $cv_{t,s}$ is the word attention score (scalar) on the $s$-th question word. Finally, $c_t$ is the textual parameter supplied to the modules at time-step $t$, containing question information needed for a sub-task.

\subsection{Neural modules with a memory stack}

\myparagraph{Module implementation.} Following the terminology in N2NMN, a neural module is a differentiable function with some internal trainable parameters, and can be used to perform a specific sub-task. For example, the question ``how many objects are right of the blue object?'' can be possibly answered by the layout \texttt{Answer[`how many'](Transform[`right'](Find[`blue']()))}, where the modules such as \texttt{Transform} are selected with module weight $w^{(t)}$ and the textual information such as `blue' is contained in the textual parameter $c_t$.

The module implementation basically follows \cite{hu2017learning}. We also simplify the implementation in \cite{hu2017learning} by merging unary answering modules (\texttt{Count}, \texttt{Exist}, \texttt{Describe}) into a single \texttt{Answer} module, and pairwise comparison (\texttt{More}, \texttt{Less}, \texttt{Equal}, \texttt{Compare}) into a single \texttt{Compare} module. Finally, we introduce a \texttt{NoOp} module that does nothing, which can be used to pad arbitrary module layouts to a maximum length $T$. Our module implementation is summarized in Table~\ref{tab:modules}. 

\begin{table}[t]
\centering
\begin{tabular}{|l|c|c|c|l|}
\hline
module & input & output & implementation details \\
name & attention & type & ($x$: image feature map, $c$: textual parameter) \\
\hline
\texttt{Find} & (none) & attention & $a_{out}=\mathrm{conv_2}\left(\mathrm{conv_1}(x) \odot W c\right)$ \\
\texttt{Transform} & $a$ & attention & $a_{out}=\mathrm{conv_2}\left(\mathrm{conv_1}(x) \odot W_1\sum(a \odot x) \odot W_2 c\right)$ \\
\texttt{And} & $a_1, a_2$ & attention & $a_{out} = \mathrm{minimum}(a_1, a_2)$ \\
\texttt{Or} & $a_1, a_2$ & attention & $a_{out} = \mathrm{maximum}(a_1, a_2)$ \\
\texttt{Filter} & $a$ & attention & $a_{out} = \mathtt{And}(a, \mathtt{Find}())$, i.e. reusing \texttt{Find} and \texttt{And} \\
\texttt{Scene} & (none) & attention & $a_{out}=\mathrm{conv_1}(x)$ \\
\texttt{Answer} & $a$ & answer & $y=W_1^T \left(W_2\sum(a \odot x) \odot W_3 c\right)$ \\
\texttt{Compare} & $a_1, a_2$ & answer & $y=W_1^T \left(W_2\sum(a_1 \odot x) \odot W_3\sum(a_2 \odot x) \odot W_4 c\right)$ \\
\texttt{NoOp} & (none) & (none) & (does nothing) \\
\hline
\end{tabular}
~\\~\\
\caption{Neural modules used in our model. The modules take image attention maps as inputs, and output either a new image attention $a_{out}$ or a score vector $y$ over all possible answers ($\odot$ is elementwise multiplication; $\sum$ is sum over spatial dimensions).}
\label{tab:modules}
\vspace{-1.5em}
\end{table}

\myparagraph{Differentiable memory stack.} In our model, different modules may take different numbers of input, and the model sometimes needs to take what it currently sees and compare it with what it has previously seen before. This is typical in tree-structured layouts, such as \texttt{Compare(Find(), Transform(Find()))}. To handle tree-structured layouts, the model needs to have a memory to remember the outputs from the previous reasoning time-steps. Similar to Memory Networks \cite{sukhbaatar2015end}, we provide a differentiable memory pool to store and retrieve the intermediate outputs. However, to encourage compositional behavior, we restrict our memory pool to be a Last-In-First-Out (LIFO) stack similar to \cite{grefenstette2015learning}. The LIFO behavior encourages the neural modules to work like functions in a computer program, allowing only arguments and returned values to be passed between the modules, without arbitrary memory modification.

Our memory stack can be used to store vectors with fixed dimensions. It consists of a length-$L$ memory array $A=\{A_i\}_{i=1}^L$ (where $L$ is the stack length) and a stack-top pointer $p$, implemented as a $L$-dimensional one-hot vector. The stack $(A, p)$ implements differentiable push and pop operations as follows. Pushing a new vector $z$ into stack $(A, p)$ is done via pointer increment as $p := \text{1d\_conv}(p, [0, 0, 1])$ followed by value writing as $A_i := A_i \cdot (1 - p_i) + z \cdot p_i$, $\text{for each } i = 1,...,L$. Similarly, popping the current stack-top vector $z$ from stack $(A, p)$ is done via value reading as $z := \sum_{i=1}^L A_i \cdot p_i$ followed by pointer decrement as $p := \text{1d\_conv}(p, [1, 0, 0])$.
Here $A_i$ is the vector at stack depth $i$ in $A$. In both push and pop operations, the one-hot stack pointer $p$ is incremented or decremented using 1-d convolution.

In our model, we use the above memory stack to store the $H\times W$ dimensional image attention maps, where $H$ and $W$ are the height and the width of the image feature map. Using the memory stack, each module first pops from the stack to obtain input image attentions, and then pushes its result back to the stack. For example, in tree-like layouts such as \texttt{Compare(Find(), Transform(Find()))}, the \texttt{Find} module pushes its localization result into the stack, the \texttt{Transform} module pops one image attention map from the stack and pushes back the transformed attention, and the \texttt{Compare} module pops two image attention maps and uses them to predict the answer.

\subsection{Soft program execution}
Our model performs continuous selection of module layout through the soft module weights $w^{(t)}$. At each time step $t$, we execute all the modules in our module list $M$ (shown in Table~\ref{tab:modules}), and perform a weighted average of their results with respect to the weights $w^{(t)}$ predicted by the layout controller. Specifically, the resulting memory stacks from the execution of each module are weighted-averaged with respect to $w_m^{(t)}$ to produce a single updated memory stack.

At time step $t=0$, we initialize the memory stack $(A, p)$ with uniform image attention and set stack the pointer $p$ to point at the bottom of the stack (one-hot vector with 1 in the 1st dimension). Then, at each time step $t$, for every module $m \in M$ we execute it on the current memory stack $(A^{(t)}, p^{(t)})$. During execution, each module $m$ may pop from the stack and push back its results, producing an updated stack $(A^{(t)}_m, p^{(t)}_m)$ as $\left(A^{(t)}_m, p^{(t)}_m\right) = \mathrm{run\_module}\left(m, A^{(t)}, p^{(t)}\right)$, $\text{for each } m \in M$. We average the resulting new stack from each module according to its weight $w_m^{(t)}$ as $A^{(t+1)} = \sum_{m \in M} A^{(t)}_m \cdot w_m^{(t)}$, and then sharpen the stack pointer with a softmax operation to keep it as a (nearly) one-hot vector as $p^{(t+1)} = \mathrm{softmax}\left(\sum_{m \in M} p^{(t)}_m \cdot w_m^{(t)}\right)$.

\myparagraph{Final output.} We apply this model to both the Visual Question Answering (VQA) task and the Referential Expressions Grounding (REF) task. To obtain the answer in the VQA task, we collect the output answer logits (i.e. scores) in all time-steps from those modules that have answer outputs (\texttt{Answer} and \texttt{Compare} in Table~\ref{tab:modules}), and accumulate them with respect to their module weights as $y = \sum_{t=0}^{T-1} \sum_{m \in M_\mathrm{ans}} y_m^{(t)} w_m^{(t)}$
where $M_\mathrm{ans}$ contains \texttt{Answer} and \texttt{Compare}.

To output grounding results in the REF task, we take the image-attention map at the top of the final stack at $t=T$, and extract attended image features from this attention map. Then, a linear layer is applied on the attended image feature to predict the bounding box offsets from the feature grid location.

\subsection{Training}

Unlike previous modular approaches N2NMN \cite{hu2017learning}, PG+EE \cite{johnson2017inferring} and TbD \cite{mascharka2018transparency}, our model does not require expert layouts to achieve good performance. When such expert layout supervision is available, our model can also utilize it by supervising the soft module weights $w_{(t)}$ with a cross-entropy loss to match the expert's module choice.
But as the entire network is fully differentiable, it can be trained effectively without reinforcement learning, from task supervision alone, in the absence of expert layout supervision.

For VQA, we train with softmax cross entropy loss on the final answer scores $y$. For REF, we map the center of the ground-truth bounding box to a location on the feature grid. Then we train with a softmax cross entropy loss on the final image attention map to put all the attention on the ground-truth feature grid, and a bounding box regression loss on the bounding box offsets to match the ground-truth box. We train with the Adam optimizer with $10^{-4}$ learning rate. 
Our code and data are available at \url{http://ronghanghu.com/snmn/}.

\section{Experiments}

We evaluate our model on the Visual Question Answering (VQA) task on the large-scale CLEVR dataest \cite{johnson2017clevr}. The dataset consists of 70000, 15000 and 15000 images for training, validation and test, and each image is associated with 10 questions. The images in the dataset are rendered from a graphics engine, and the questions are synthesized with complex reasoning procedure.

To evaluate our model on the Referential Expression Grounding (REF) task \cite{rohrbach2016grounding}, we build a new \textit{CLEVR-Ref} dataset with images and referential expressions in CLEVR style using the code base of \cite{johnson2017clevr}. Our new CLEVR-Ref dataset has the same scale as the original CLEVR dataset for VQA, but contains referential expressions instead of questions. Each referential expression refers to a unique object in the image, and the model is required to ground (i.e. localize) the corresponding object with a bounding box. The grounded bounding box is considered correct if it overlaps with the ground-truth bounding box by at least 0.5 intersection-over-union (IoU). Similar to question answering in the CLEVR dataset, the referential expressions also involve complex reasoning and relationship handling. See Figure~\ref{fig:clevr_vis} for an example of the CLEVR-Ref dataset.

\subsection{Model performance}

\begin{table}[t]
\centering
\begin{tabular}{c@{~~~~}c@{~~~~}c@{~~~~}c}
\toprule
trained on & expert layout & VQA accuracy & REF accuracy \\
\hline
VQA & yes & \textbf{96.6} & n/a \\
REF & yes & n/a & 96.0 \\
VQA+REF & yes & 96.5 & \textbf{96.2} \\
\hline
VQA & no & 93.0 & n/a \\
REF & no & n/a & 93.4 \\
VQA+REF & no & \textbf{93.9} & \textbf{95.4} \\
\bottomrule
\end{tabular}
~\\~\\
\caption{Validation accuracy on the CLEVR dataset (VQA) and the CLEVR-Ref dataset (REF). Our model simultaneously handles both tasks with high accuracy.}
\label{tab:clevr_multitask}
\vspace{-1.5em}
\end{table}

Our model aims to simultaneously handle both VQA and REF tasks, and to decompose the reasoning procedure into sub-tasks by inducing a suitable module layout on each question or referential expression.

We train our model on the CLEVR dataset for the VQA task, and the CLEVR-Ref dataset for the REF task. We experiment with training only on the VQA task, training only on the REF task, and joint training on both tasks (VQA+REF) using the loss from both tasks. To test whether our model can induce a reasonable sub-task decomposition and module layout, we experiment with both using expert layout supervision (same as in \cite{hu2017learning}) and training from scratch without expert layout. We use a ResNet-101 convnet \cite{he2016deep} pretrained on ImageNet classification to extract visual features from the image.

\begin{figure}[t]
\begin{tabular}{ccccccccc}
VQA (expert layout) & &
VQA (from scratch) & &
REF (expert layout) & &
REF (from scratch)
\\
\includegraphics[trim={0 2mm 0 3mm},clip,width=.20\textwidth,valign=T]{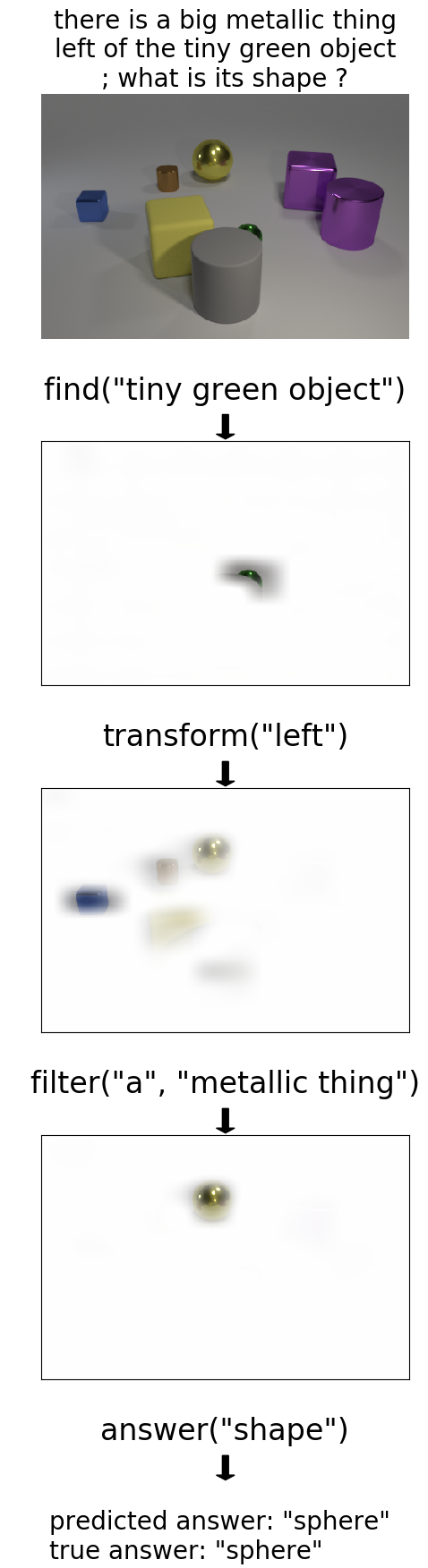} & &
\includegraphics[trim={0 2mm 0 3mm},clip,width=.2\textwidth,valign=T]{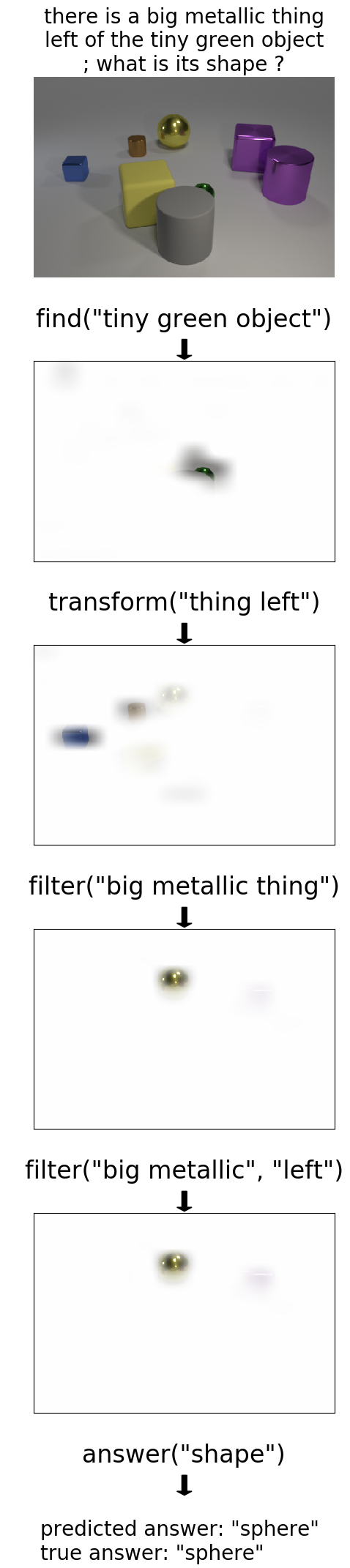} & &
\includegraphics[trim={0 2mm 0 3mm},clip,width=.20\textwidth,valign=T]{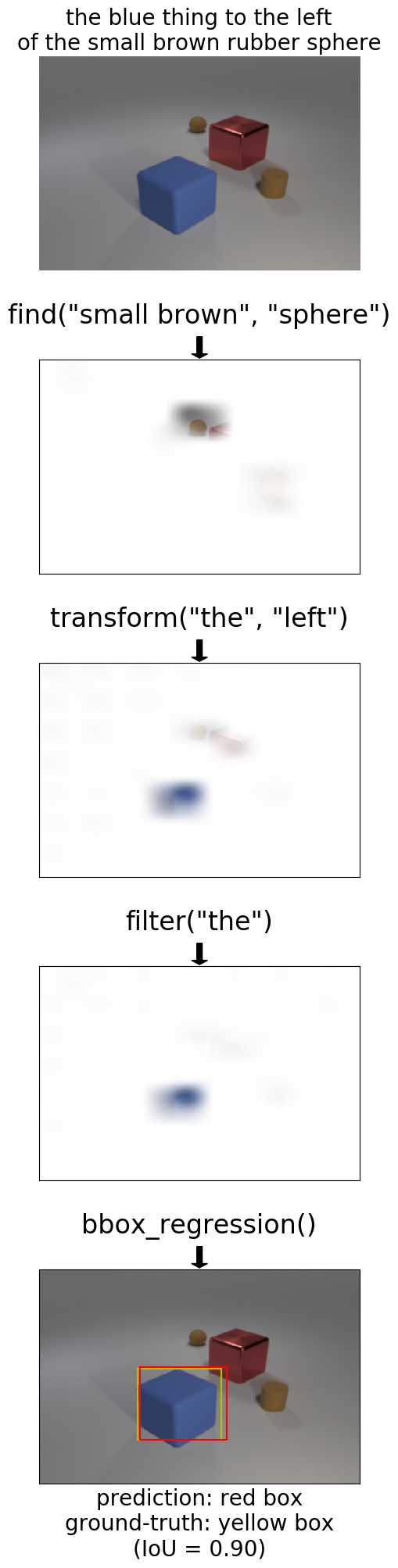} & &
\includegraphics[trim={0 2mm 0 3mm},clip,width=.194\textwidth,valign=T]{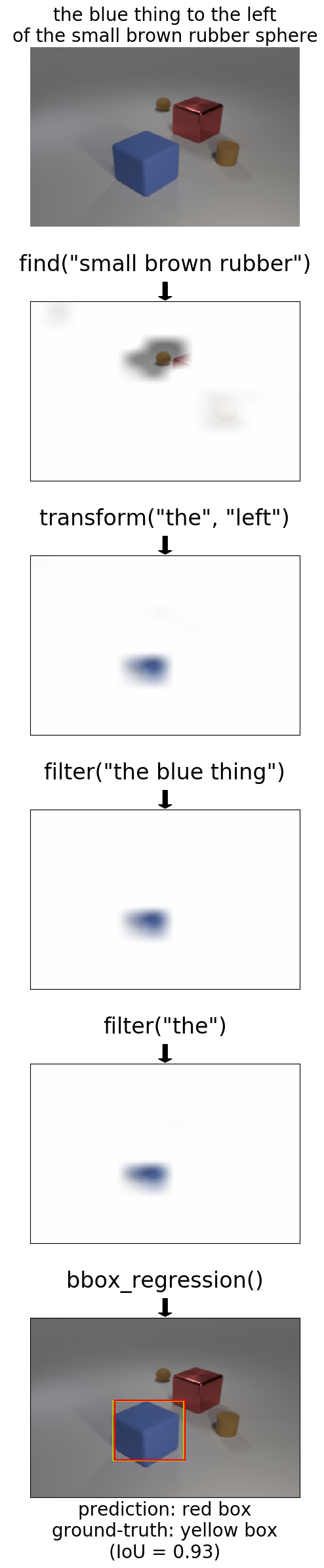} \\
\end{tabular}
\caption{Examples of our model on VQA (left) and REF (right). At each step, we visualize the module with the highest weight, the words receiving most textual attention ($cv_{t,s}$ in Sec.~\ref{sec:controller}) and the module output.}
\label{fig:clevr_vis}
\end{figure}

The results are summarized in Table~\ref{tab:clevr_multitask}. It can be seen that when training on each individual task, our model achieves over 90\% accuracy on both tasks (which is reasonably good performance), whether using expert layout supervision or not. Furthermore, joint training can lead to even higher accuracy on these two tasks (especially when not using expert layout). Our model can simultaneously handle these two tasks by exploiting the common sub-tasks in them, such as finding object and handling relationships.

\myparagraph{Sub-task decomposition and layout induction.} By comparing the bottom 3 rows (trained without using expert layout) and the top 3 rows (trained with expert layout supervision), it can be seen that although the models trained with expert layout still outperforms training from scratch, the gap between the two scenarios is relatively small. This indicates that our model can still work well without layout supervision, which is something previous modular approaches such as N2NMN \cite{hu2017learning}, PG+EE \cite{johnson2017inferring} and TbD \cite{mascharka2018transparency} could not handle.

We visualize the reasoning procedure our multi-task model on both VQA and REF task, for both with expert layout and without expert layout supervision. Figure~\ref{fig:clevr_vis} shows the module layout, the intermediate reasoning outputs and the most attended words from textual attention ($cv_{t,s}$ in Sec.~\ref{sec:controller}). It can be seen that our model can induce a reasonable decomposition of the inference procedure into sub-tasks without expert layout supervision, and it learns to share common sub-tasks such as find (localization) and transform in across the two tasks.

We note that our model learns peaky module weights after convergence. The average entropy of the learned soft module weights (which can be seen as a probability distribution) is $0.082$ when trained without layout supervision (corresponds to putting over $98\%$ weights on one module), and $7.5 \times 10^{-5}$ when trained with layout supervision (corresponds to putting over $99.99\%$ weights on one module). This shows that even without any strong supervision on module layout, our model learns to almost discretely select one module through the soft module weights at test time. Hence, our proposed framework can be regarded as a novel end-to-end differentiable training approach for modular networks.

We further experiment with test-time layout discretization by replacing the soft module weights with a one-hot argmax vector. This results in sightly lower performance on the CLEVR validation set (90.0\% when trained without layout supervision and 94.8\% with layout supervision). Considering the discrepancy between training and test time, the relatively small accuracy drop ($<4\%$) from test-time layout discretization indicates that our model works similar to previous modular networks at test time, rather than acting as a mixture of experts.

\myparagraph{Evaluation of accuracy.} We first compare the accuracy of our model on the CLEVR VQA dataset with the previous modular approaches N2NMN \cite{hu2017learning}, PG+EE \cite{johnson2017inferring} and TbD \cite{mascharka2018transparency}. N2NMN uses a layout policy to predict discrete layouts and a neural module network to answer the question. PG+EE and TbD are also modular approachs similar to N2NMN, where the program generator is similar to the layout policy, and the execution engine is essentially a neural module network. For fair comparison with previous work, we train our model on the CLEVR VQA dataset only (without using CLEVR-Ref for joint training).

The results are shown in Table~\ref{tab:clevr_compare}. It can be seen from the top 4 rows that among all the modular approaches (N2NMN, PG+EE, TbD and Ours), when layout supervision is available, our model outperforms N2NMN by a large margin, and achieves comparable performance with PG+EE while underperforms TbD by a small margin. We note that even when using expert layout, our model still uses less supervision than PG+EE or TbD as they both require fine-grained module specification (e.g. finding shape and finding color are different modules in \cite{johnson2017inferring,mascharka2018transparency} while the same module with different textual attention in our model).

The bottom 4 rows show the results without using expert layout supervision, where our model significantly outperform N2NMN. In this case, N2NMN has large performance drop while PG+EE and TbD fails to converge or cannot not be trained without layout supervision. This can be attributed to the fact that N2NMN, PG+EE and TbD all use discrete non-differentiable layout, while our model is fully differentiable and can be trained with back-propagation.

We note that the best non-modular architectures \cite{hudson2018compositional} achieve higher performance without using expert layout supervision, and compare those against modular performance on both accuracy and interpretability in Sec.~\ref{sec:exp_compare}.

\begin{table}[t]
\centering
\begin{tabular}{l@{~~~~}c@{~~~~}c}
\toprule
Method & expert layout & accuracy on CLEVR \\
\hline
N2NMN \cite{hu2017learning} & yes & 83.7 \\
PG+EE \cite{johnson2017inferring} & yes & 96.9 \\
TbD \cite{mascharka2018transparency} & yes & \textbf{99.1} \\
Ours & yes & 96.5 \\
\hline
N2NMN \cite{hu2017learning} & no & 69.0 \\
PG+EE \cite{johnson2017inferring} & no & (does not converge) \\
TbD \cite{mascharka2018transparency} & no & (not supported) \\
Ours & no & \textbf{93.0} \\
\bottomrule
\end{tabular}
~\\~\\
\caption{Comparison of our model and other modular approaches on the CLEVR dataset for VQA. Our model achieves the best accuracy when not relying on expert layout, while N2NMN has significant performance drop in this case. The best non-modular architectures (e.g., \cite{hudson2018compositional}) do achieve higher performance; we compare those against modular performance on both accuracy and interpretability in Sec.~\ref{sec:exp_compare}.}
\label{tab:clevr_compare}
\vspace{-0.5em}
%
\centering
\begin{tabular}{l@{~~~~~}c@{~~~~~}c@{~~~~~}c}
\toprule
Method & expert layout & accuracy on VQAv1 & accuracy on VQAv2 \\
\midrule
N2NMN \cite{hu2017learning} & yes & 64.9 & 63.3 \\
ours & no & 65.5 & \textbf{64.1} \\
ours & yes & \textbf{66.0} & 64.0 \\
\bottomrule
\end{tabular}
~\\~\\
\caption{Single-model accuracy of our method and N2NMN \cite{hu2017learning} on both VQAv1 \cite{antol2015vqa} and VQAv2 \cite{goyal2017making}
datasets, using the same experimental settings (e.g. visual features).}
\label{tab:vqa_dataset}
\vspace{-1.5em}
\end{table}

\myparagraph{Results on real-image VQA datasets.}
We also evaluate our method on real-image visual question answering datasets and compare with N2NMN \cite{hu2017learning}. We run our approach on both VQAv1 and VQAv2 datasets \cite{antol2015vqa,goyal2017making} following the same settings (e.g. using ResNet-152 image features and single model at test time without ensemble) as in \cite{hu2017learning}, where the results are in Table~\ref{tab:vqa_dataset}. Although the question answering task in these datasets focuses more on visual recognition than on compositional reasoning, our method still outperforms \cite{hu2017learning} even without expert layout supervision (the expert layouts are obtained by a syntactic parser).

\subsection{Model interpretability}
\label{sec:exp_compare}

\myparagraph{Evaluation of interpretability.} It is often suggested in existing works \cite{hu2017learning,johnson2017inferring,mascharka2018transparency} that modular networks can be more interpretable to humans compared to holistic models. However, there is a lack of human studies in these works to support this claim.
In this section, we evaluate how well the user can understand the internal reasoning process within our model, and compare it with MAC \cite{hudson2018compositional}.\footnote{In our comparison, we use a fixed number of 12 steps for MAC (following the default setting in \cite{hudson2018compositional}), which is longer than the average number of steps in our Stack-NMN model.} 
We compare to MAC because it is a state-of-the-art holistic model that also performs multi-step sequential reasoning and has image and textual attention at each time-step, while other models (e.g., FiLM \cite{perez2018film} and Relation Net \cite{santoro2017simple}) have lower performance and do not have any image or textual attention to visualize. MAC is a multi-step recurrent structure with a control unit and a reading-writing unit. Similar to our model, it also attend to text and image in each reasoning step. But unlike our model, there is not explicit modular structure in MAC.

Here, we investigate two distinct, but related questions: does modular structure improve humans' \emph{subjective perceptions} of model 
interpretability, and does this structure allow users to form
\emph{truthful beliefs} about model behavior? To this end,
we present two different sets of experiments (subjective understanding and forward prediction) with human 
evaluators.
With respect to the taxonomy of interpretability evaluations presented in \cite{doshi2017towards}, these are both ``human-grounded'' metrics aimed at testing ``general notions of the quality of an explanation''.

In the \textbf{subjective understanding} evaluation, we visualize model's intermediate outputs such as the image attention and textual attention at each step, and we also show the model's final prediction. The visualizations can be seen in Figure~\ref{fig:clevr_vis}. Then the human evaluators are asked to judge how well they can understand the internal reasoning process, or whether it clear to the user what the model is doing at each step. Each example is rated on a 4-point Likert scale (\textit{clear}, \textit{mostly clear}, \textit{somewhat unclear} and \textit{unclear}) corresponding to numerical scores of 4, 3, 2 and 1. 
The averaged scores and the percentage of each choice are shown in Table~\ref{tab:clevr_human_study}, where it can be seen that our model has higher subjective understanding scores than MAC \cite{hudson2018compositional} and is much more often rated as ``clear'' in both cases (using or not using expert layout supervision). This shows that the users can more clearly understand the reasoning procedure in our model.

In the \textbf{forward prediction} evaluation, we investigate whether humans can predict the model's answer and detect its failure based on these visualizations. We split the test set into half correct and half incorrect model predictions, and the final answer output is not shown, so that human baseline performance should be chance or 50\%. Our hypothesis is that if humans can predict whether the model succeed or fail better than chance, they understand something about the model's decision process. In Table~\ref{tab:clevr_human_study}, we show the human accuracy on this task along with 95\% confidence interval. It can be seen that our model allows them to predict whether the model will get the correct answer or fail consistently higher than chance when trained without expert layout supervision. We also notice that when using supervision from expert layout, our model does worse at human prediction of model's failure. We suspect it is because predicting the answer requires human to understand how the model works. When supervising the layout, the model may overfit to the expert layout, at the expense of predictability. It may output an ``intuitive'' layout by mimicking the training data, but that layout may not actually be how it is solving the problem. On the other hand, the unsupervised model is not being forced to predict any particular layouts to minimize loss, so its layouts may be more directed at minimizing the answer loss.

Finally, we compare our model with MAC on VQA accuracy in Table~\ref{tab:clevr_human_study}. Our model underperforms the state-of-the-art MAC in terms of VQA accuracy. However, our model is more interpretable to a human user. This is in line with the intuition that there may be an accuracy-explainability tradeoff, e.g., linear models are less accurate but more interpretable than non-linear models. However, our model greatly reduces the accuracy gap with the top performing models, without requiring expert layout supervision at training time.

\begin{table}[t]
\centering
\includegraphics[width=0.8\textwidth,valign=T]{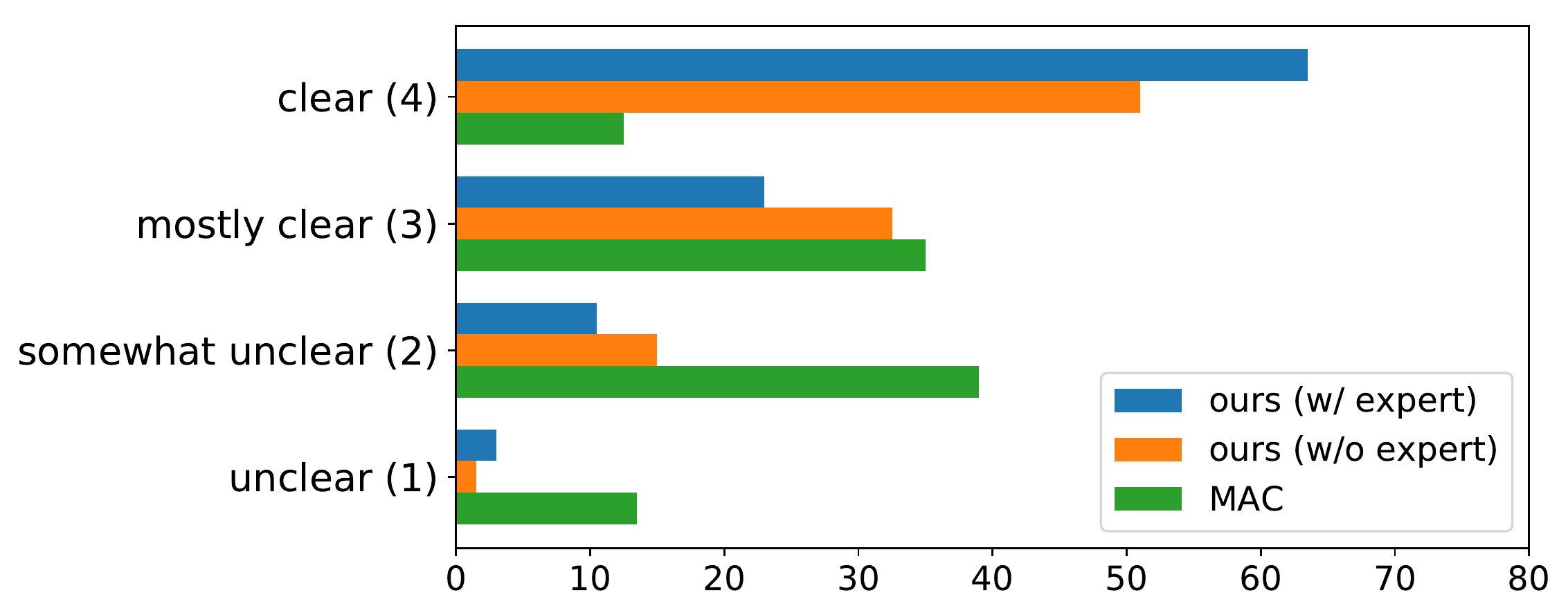} \\
percentage of each choice (\textit{clear}, \textit{mostly clear}, \textit{somewhat unclear} and \textit{unclear}) \\
\begin{tabular}[t]{l@{~~}c@{~~~}c@{~~~}c@{~~}c}
\toprule
& expert & subjective & forward prediction (failure detection) & VQA \\
Method & layout & understanding & accuracy $\pm$ 95\% confidence interval & accuracy \\
\hline
Ours & yes & \textbf{3.47} & 0.545 $\pm$ 0.069 & 96.5 \\
Ours & no & 3.33 & \textbf{0.625 $\pm$ 0.067} & 93.0 \\
MAC \cite{hudson2018compositional} & n/a & 2.46 & 0.565 $\pm$ 0.069 & \textbf{98.9} \\
\bottomrule
\end{tabular}
~\\~\\
\caption{Human evaluation of our model and the state-of-the-art non-modular MAC model \cite{hudson2018compositional}. Based on the models' intermediate outputs, the evaluators are asked to (a) judge how clearly they can understand the reasoning steps performed by these models on a 4-point scale (i.e. subjective understanding) and (b) do forward prediction (failure detection) and decide whether the model fails without seeing the final output answer. The results show that our model is more interpretable to human users. However, our model underperforms the non-modular MAC approach in VQA accuracy, which is in line with the intuition that there may be an accuracy-explainability tradeoff.}
\label{tab:clevr_human_study}
\vspace{-1.5em}
\end{table}

\section{Conclusion}

In this paper, we have proposed a novel model for visual question answering and referential expression grounding.
We demonstrate that our model simultaneously addresses both tasks by exploiting the intuition that related tasks should share common sub-tasks, and sharing a common set of neural modules between tasks. Compared with previous modular approaches, our model induces a decomposition of the inference procedure into sub-tasks while not requiring expert layout supervision. The proposed model can explain its reasoning steps with a sequence of soft module choices, image attentions, and textual attentions. Experimental evaluation found that these explanations produced better understanding in human users with respect to both subjective and objective evaluations, even in the absence of human-provided explanations at training time.

\noindent\textbf{Acknowledgements.} This work was partially supported by
US DoD and DARPA XAI and D3M, and the Berkeley Artificial Intelligence Research (BAIR) Lab.

\clearpage
\bibliographystyle{splncs04}
\bibliography{one4all}

\clearpage
\appendix

\begin{tabular}{c}
\large \textbf{Explainable Neural Computation via Stack Neural} \\
\large \textbf{Module Networks} (Supplementary Material)
\end{tabular}

\section{Visualization of our model and MAC}

\begin{figure}[h]
\centering
\includegraphics[width=\textwidth]{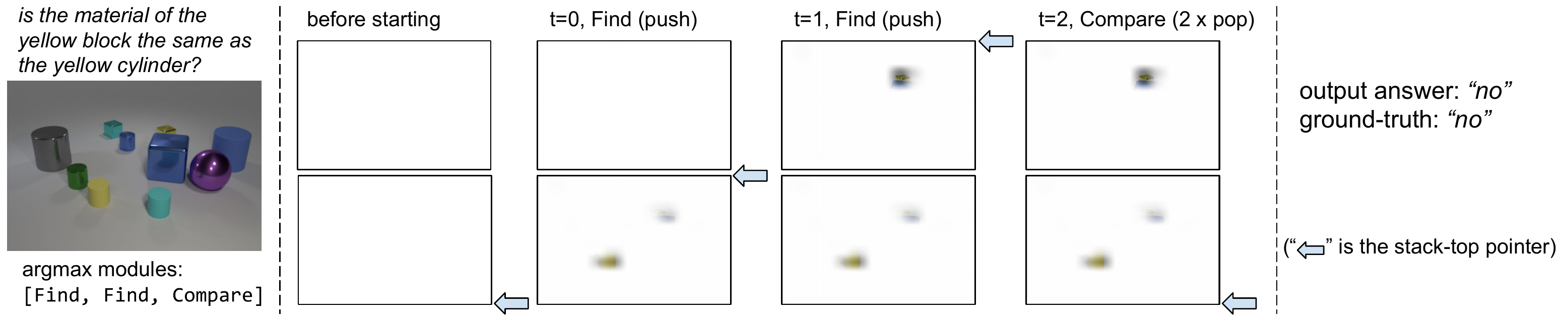}
\caption{An illustration example of our differentiable memory stack.}
\label{fig:supp_stack}
\end{figure}

The differentiable memory stack in our Stack-NMN model is a Last-In-First-Out (LIFO) data structure. We illustrate how our memory stack works with a visualized example in Figure \ref{fig:supp_stack}.

\begin{figure}[t]
\center
\footnotesize
\begin{tabular}{c|c|cc}
ours with & ours without & \multicolumn{2}{c}{MAC \cite{hudson2018compositional} with 12 steps} \\
expert layout & expert layout & Step 1 to 6 & Step 7 to 12 \\
\hline
\includegraphics[trim={0 0 0 0},clip,scale=.18,valign=T]{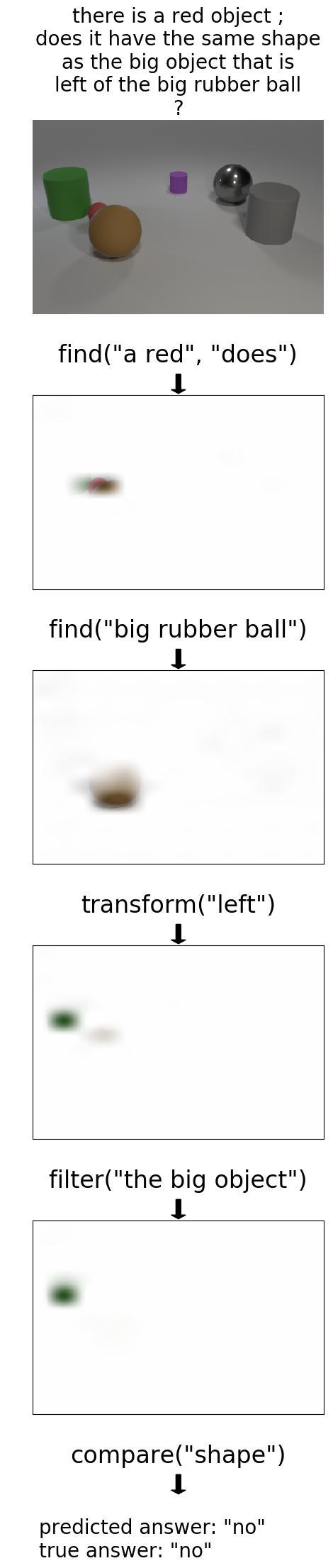} &
\includegraphics[trim={0 0 0 0},clip,scale=.18,valign=T]{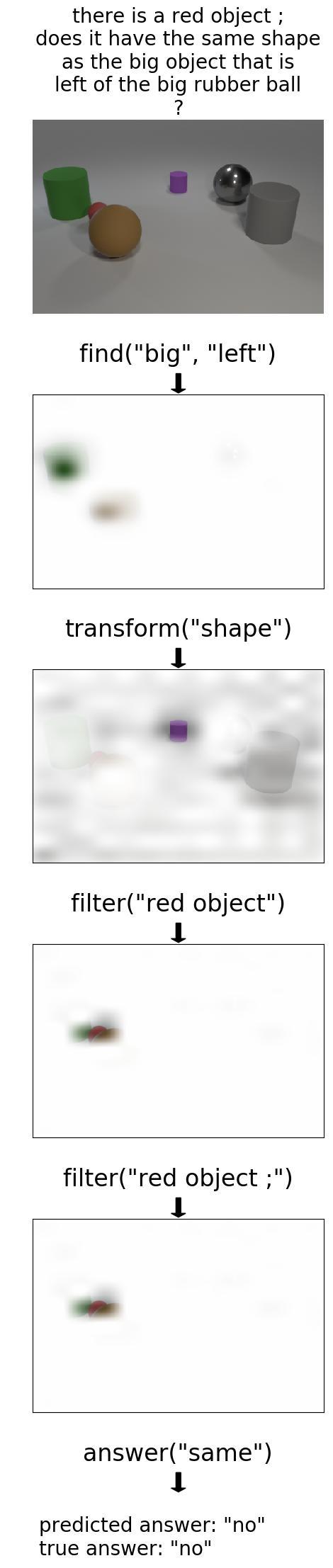} &
\includegraphics[trim={0 0 0 0},clip,scale=.18,valign=T]{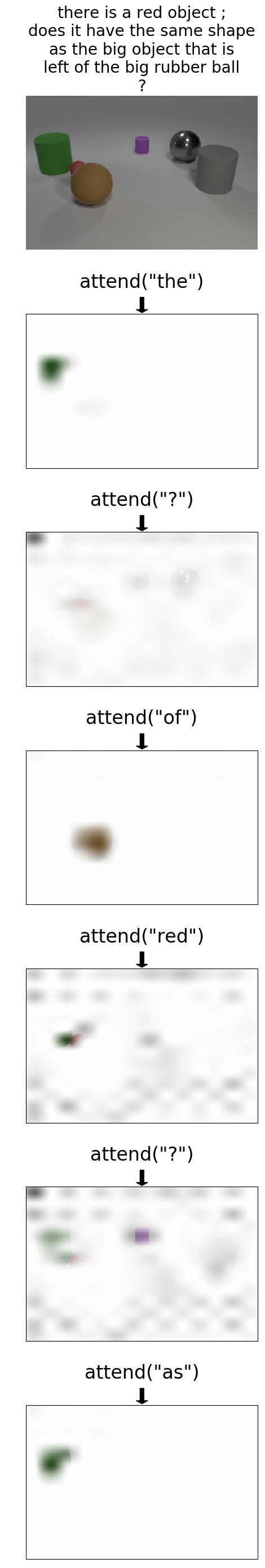} &
\includegraphics[trim={0 0 0 0},clip,scale=.18,valign=T]{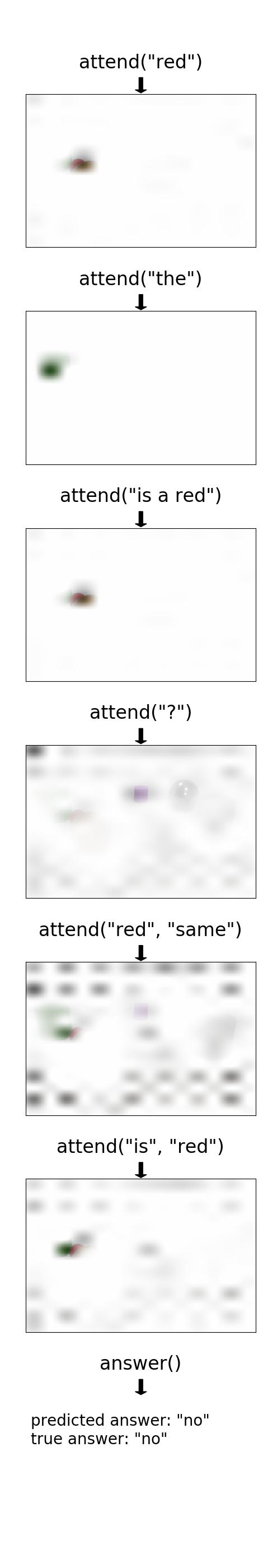} \\
\end{tabular}
\caption{Examples from of our model and the MAC model \cite{hudson2018compositional}. The MAC model has 12 reasoning steps. We visualize the per-step image attentions and textual attentions (showing the words with the most attention), and the selected module (the one with the highest weight) in our model.}
\label{fig:clevr_vis_supp}
\end{figure}

In Figure \ref{fig:clevr_vis_supp}, we show additional visualized examples of the reasoning steps in our model (trained either with or without expert layout supervision). We also use a similar visualization to show the per-step image attentions and the textual attentions from the MAC model. The MAC model \cite{hudson2018compositional} is similar to our model as it also have multiple steps (12 steps are used in \cite{hudson2018compositional}) and involves textual attention and image attention in each step. However, unlike our model, it does not have a modular structure, and uses a homogeneous unit to handle all kinds of sub-tasks in reasoning.

\section{Details of the human evaluation on interpretability}

\begin{figure}[t]
\center
\includegraphics[height=.3\textheight]{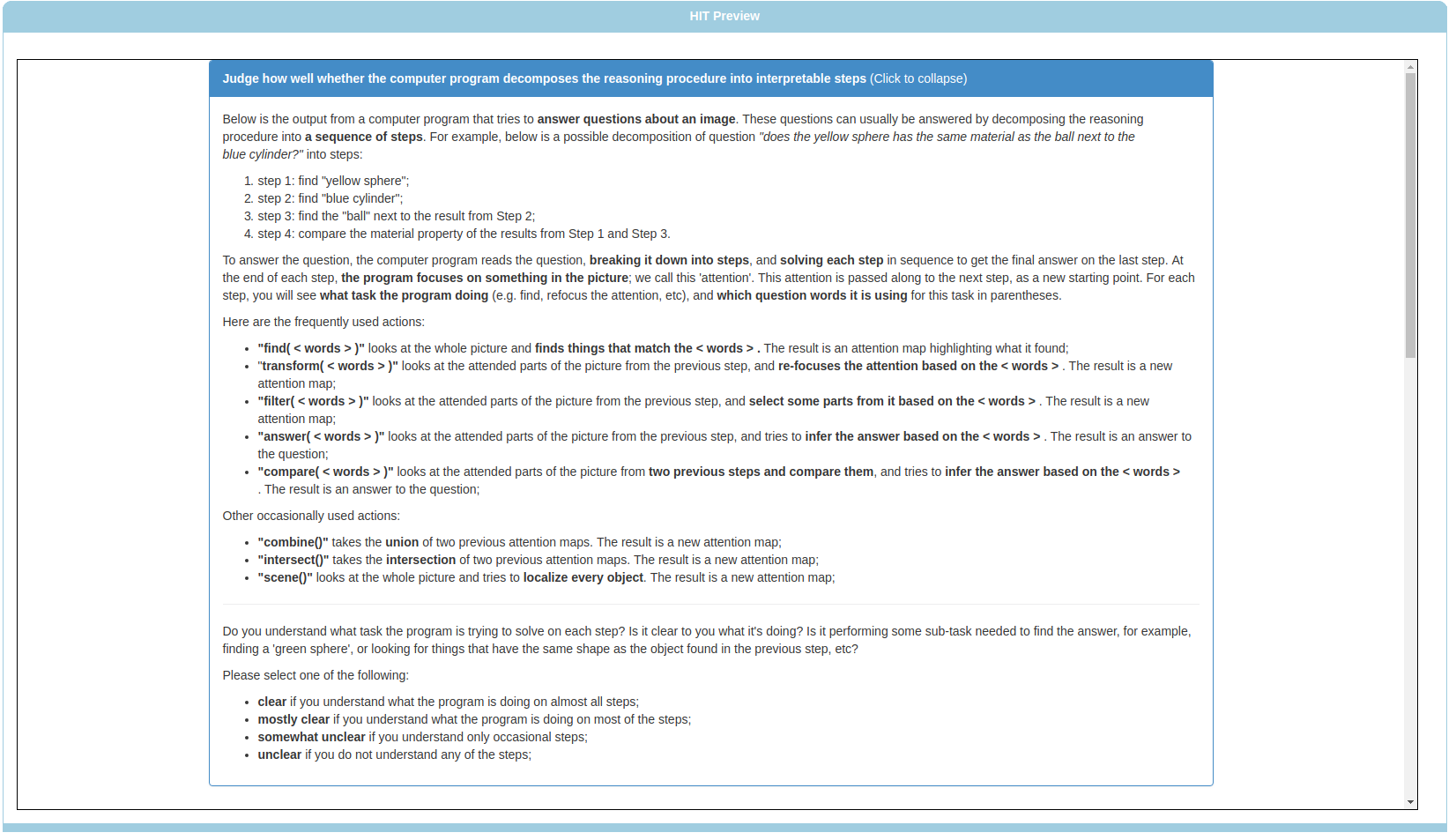} \\
\includegraphics[height=.3\textheight]{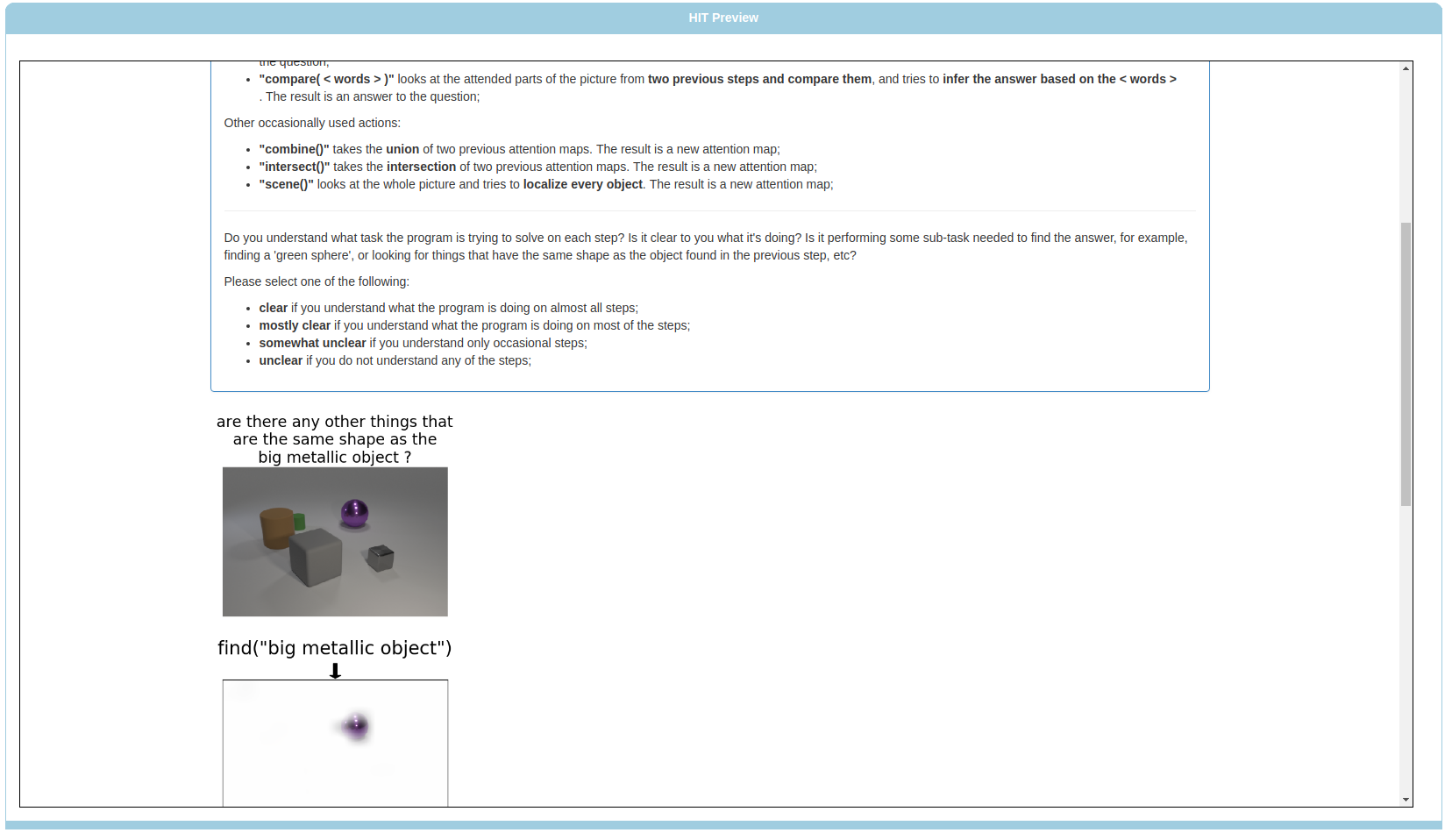} \\
\includegraphics[height=.3\textheight]{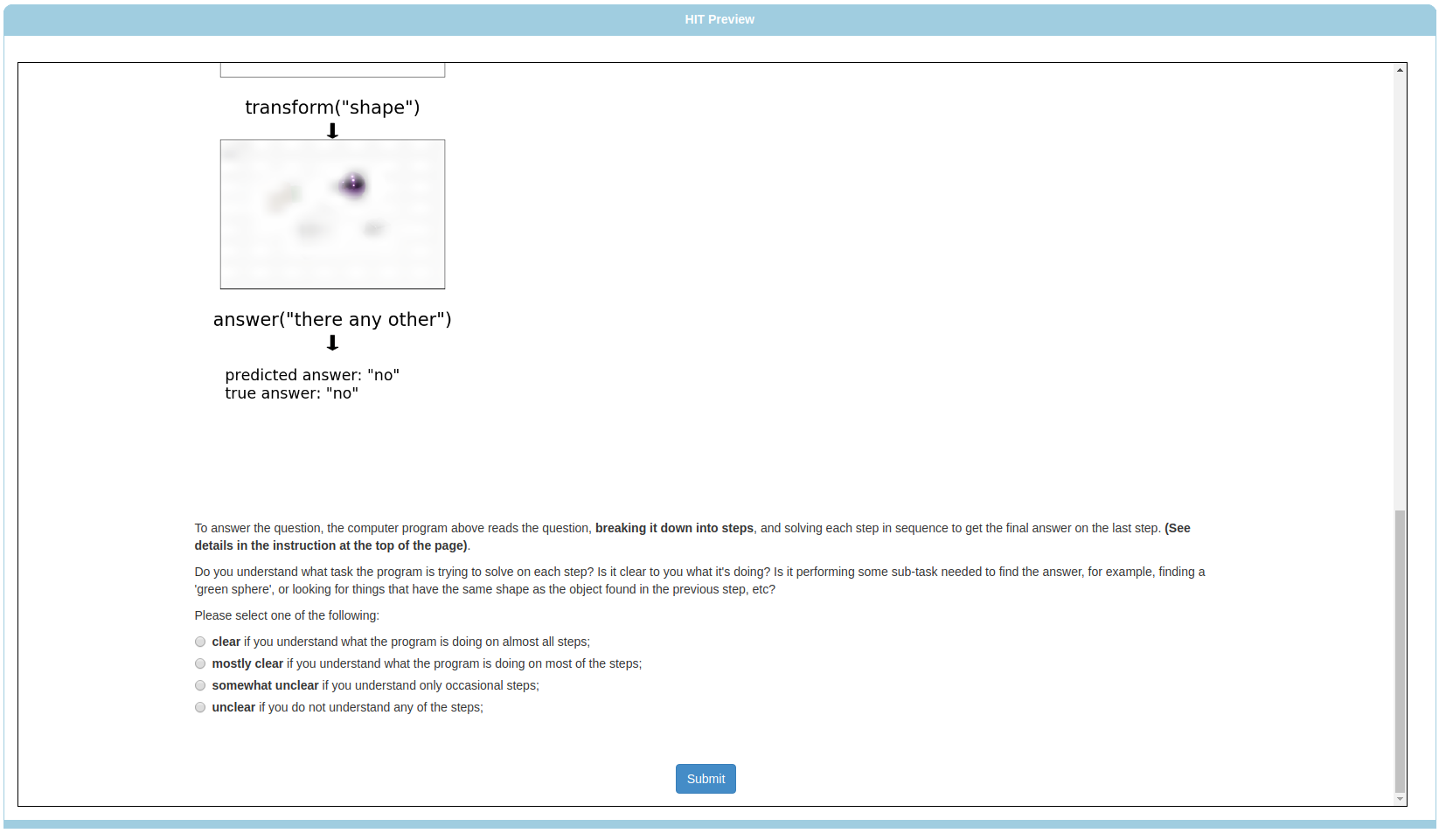}
\caption{The evaluation interface for \textit{subjective understanding} on our model, deployed on Amazon Mechanical Turk (AMT).}
\label{fig:amt_subjective_understanding_ours}
\end{figure}

\begin{figure}[t]
\center
\includegraphics[height=.3\textheight]{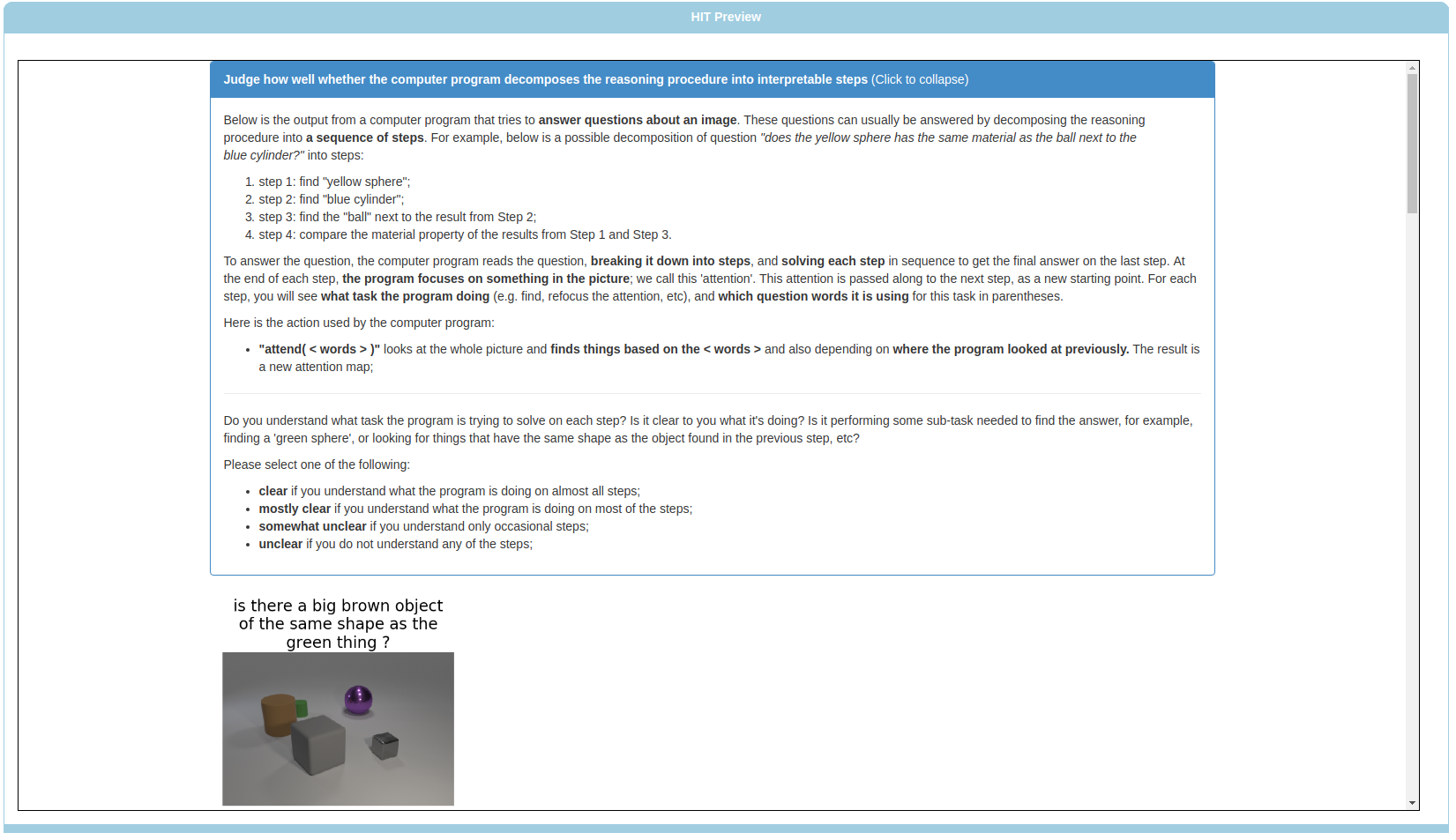} \\
\includegraphics[height=.3\textheight]{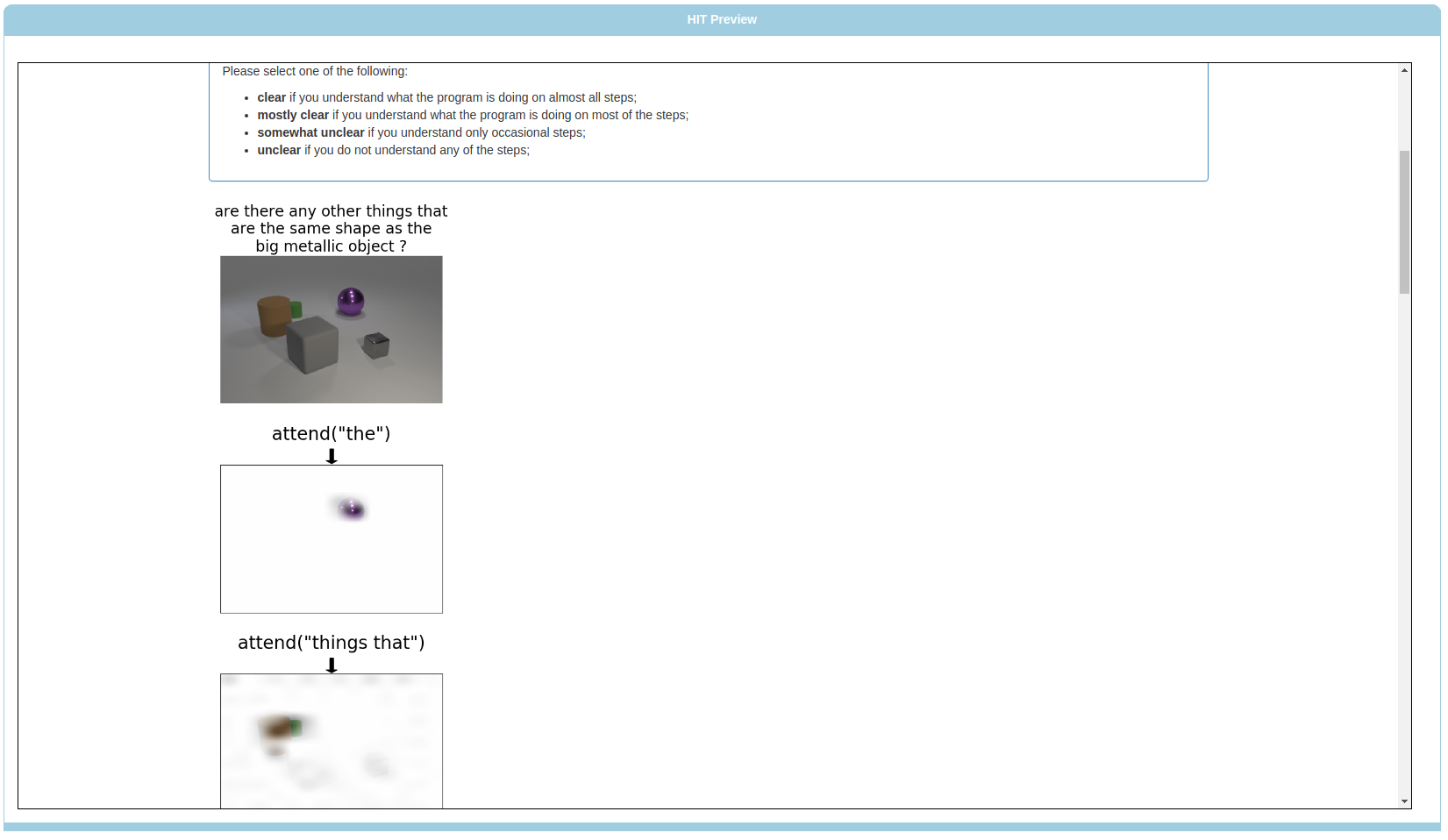} \\
\includegraphics[height=.3\textheight]{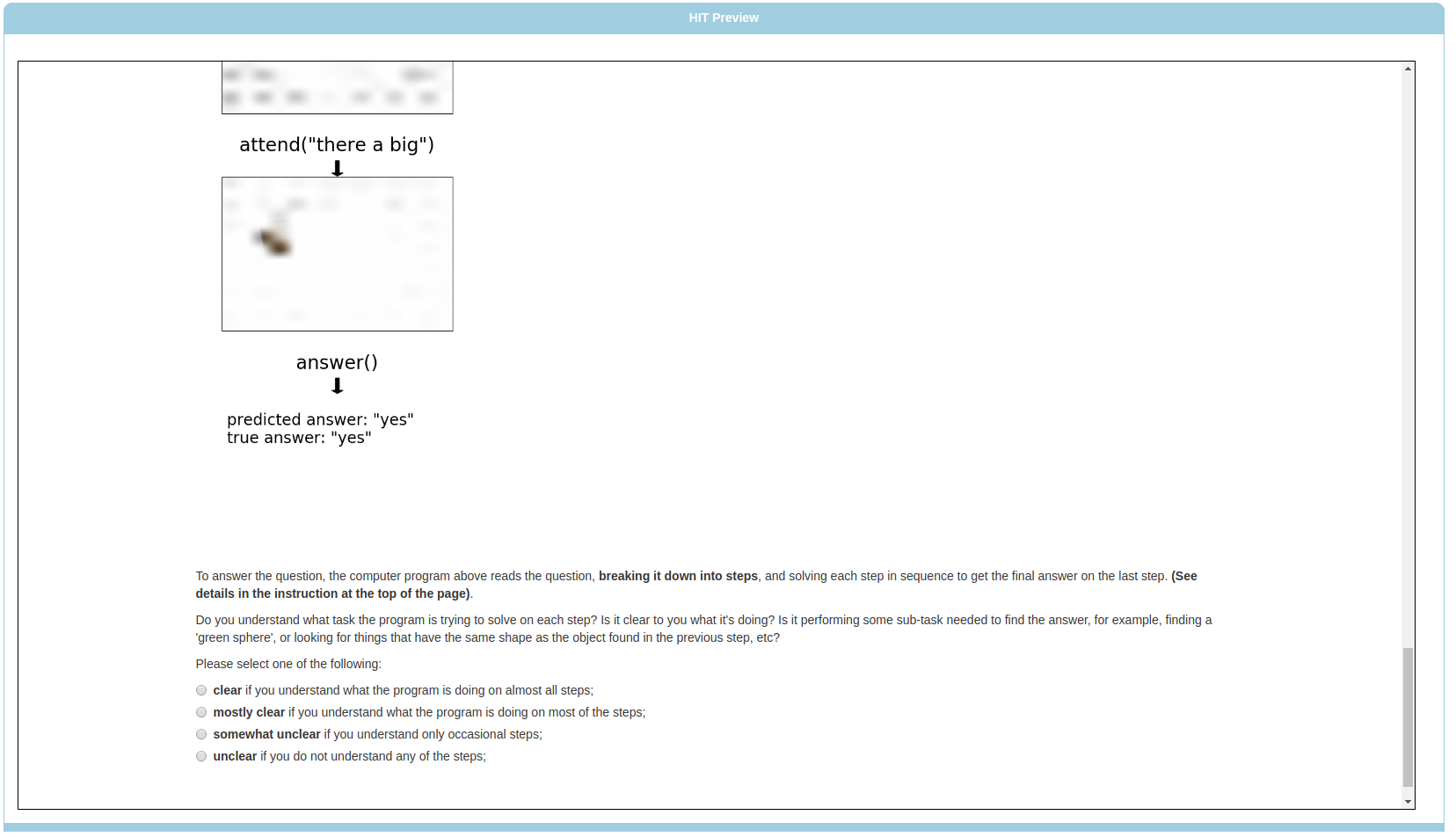}
\caption{The evaluation interface for \textit{subjective understanding} on MAC \cite{hudson2018compositional} model, deployed on Amazon Mechanical Turk (AMT).}
\label{fig:amt_subjective_understanding_mac}
\end{figure}

\begin{figure}[t]
\center
\includegraphics[height=.3\textheight]{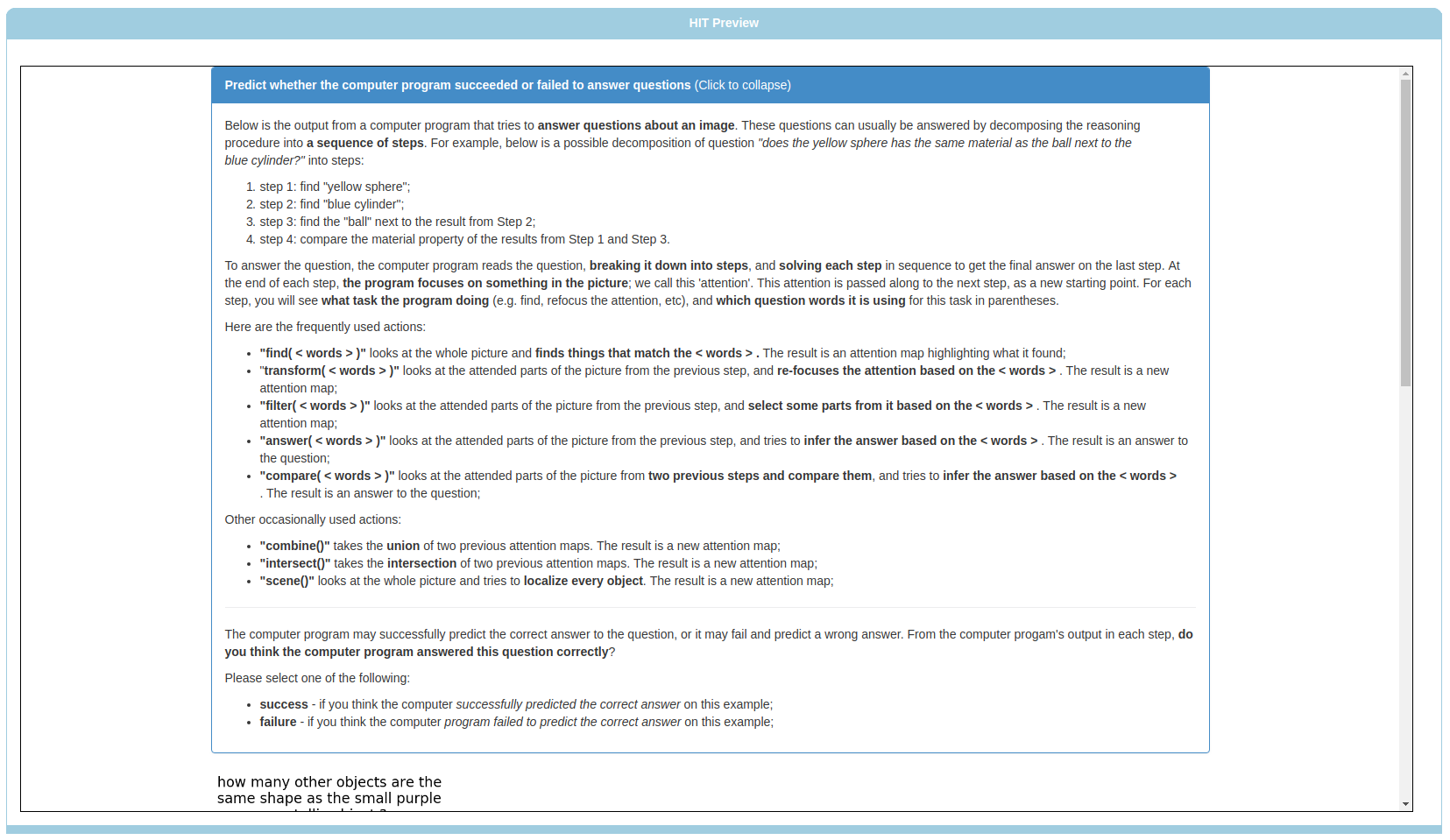} \\
\includegraphics[height=.3\textheight]{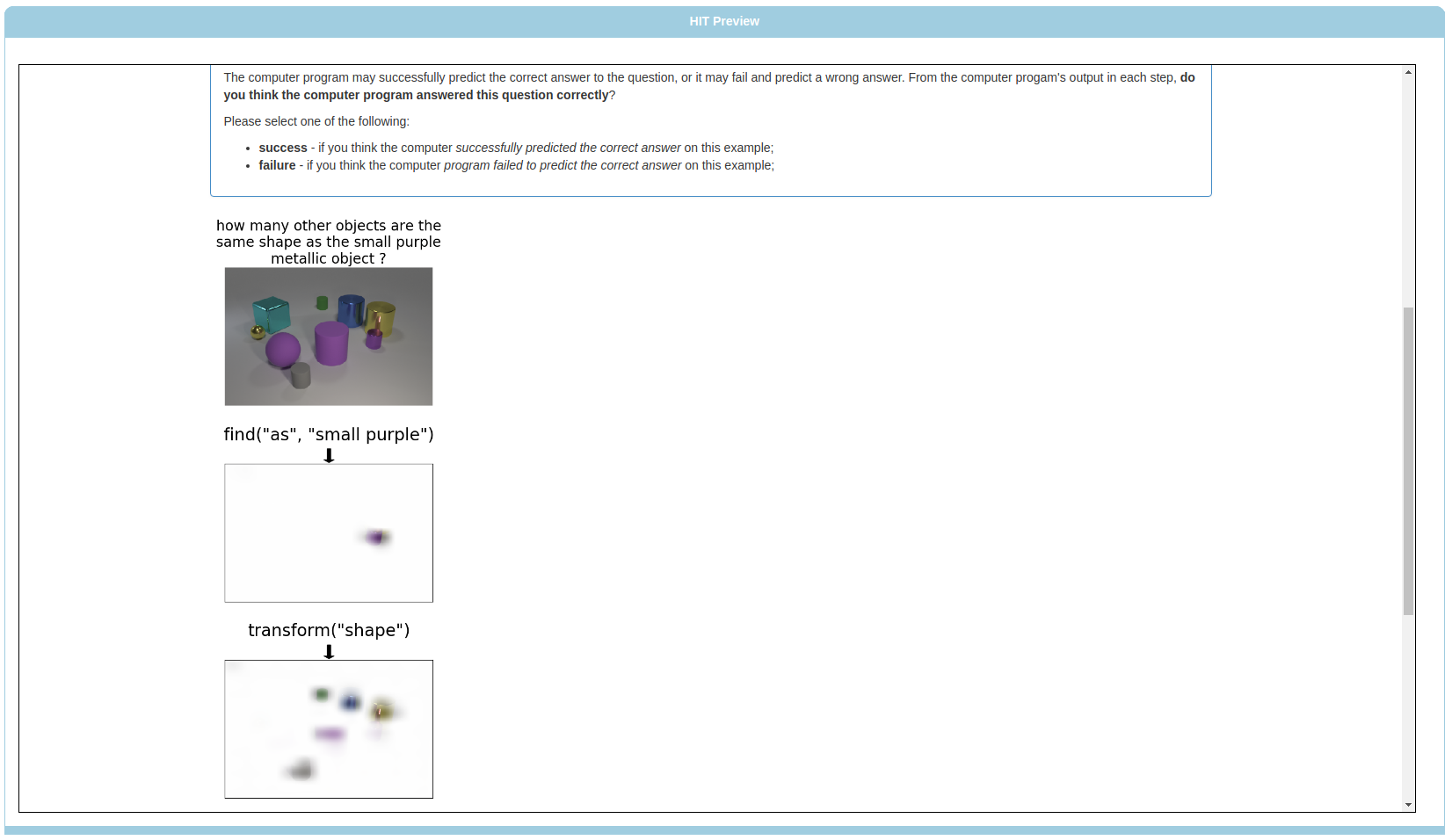} \\
\includegraphics[height=.3\textheight]{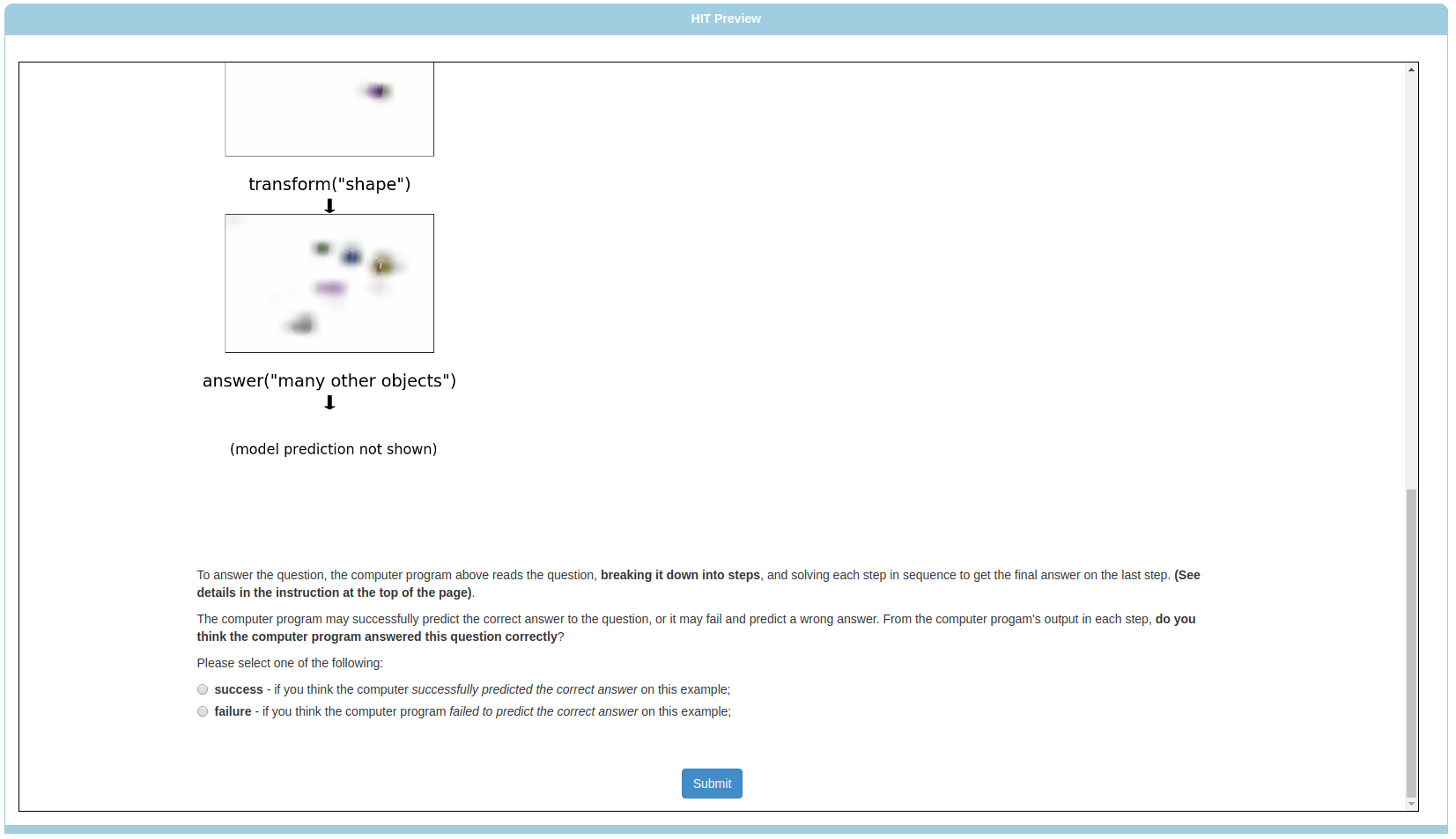}
\caption{The evaluation interface for \textit{forward prediction (failure detection)} on our model, deployed on Amazon Mechanical Turk (AMT).}
\label{fig:amt_failure_detection_ours}
\end{figure}

\begin{figure}[t]
\center
\includegraphics[height=.3\textheight]{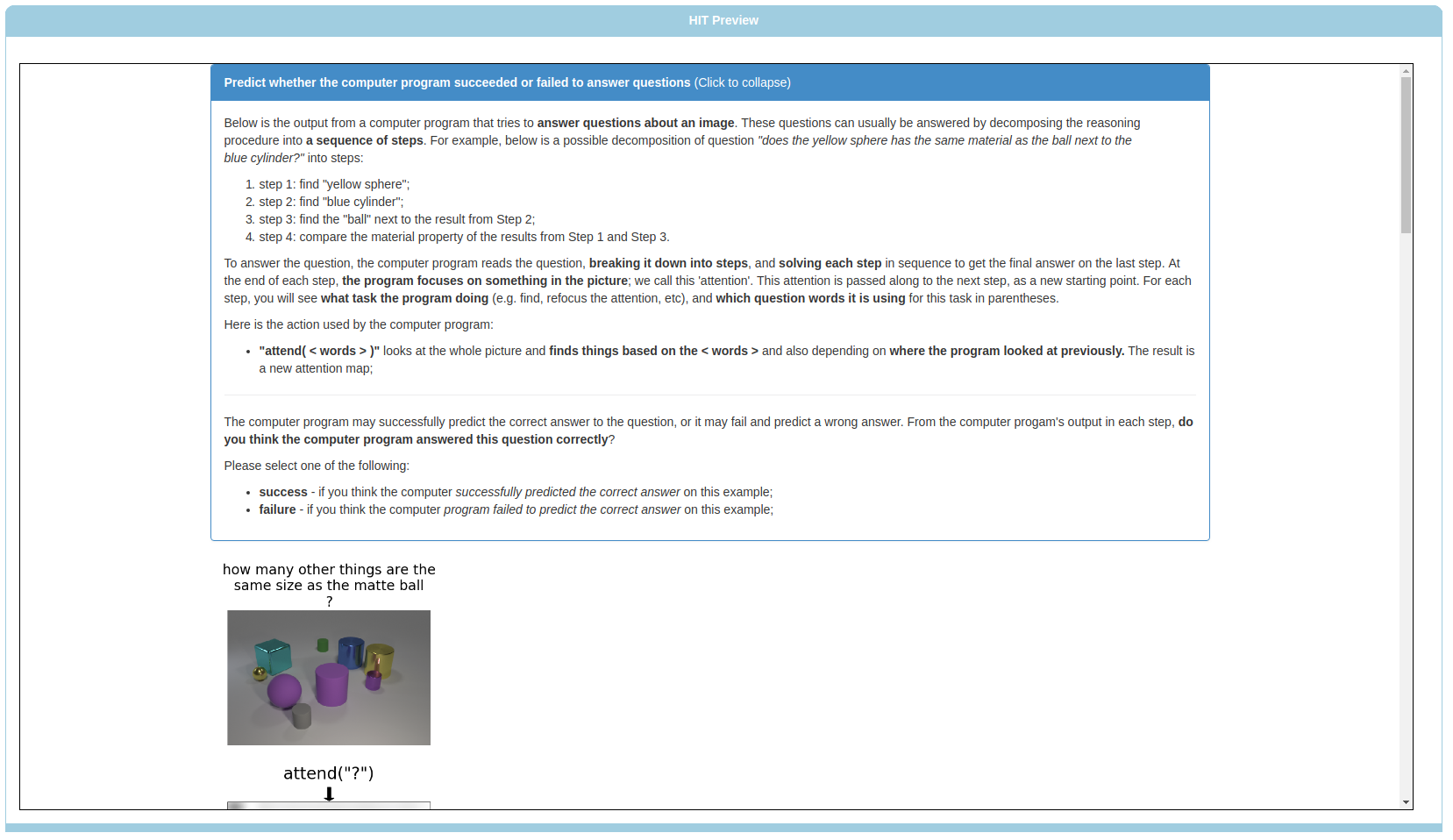} \\
\includegraphics[height=.3\textheight]{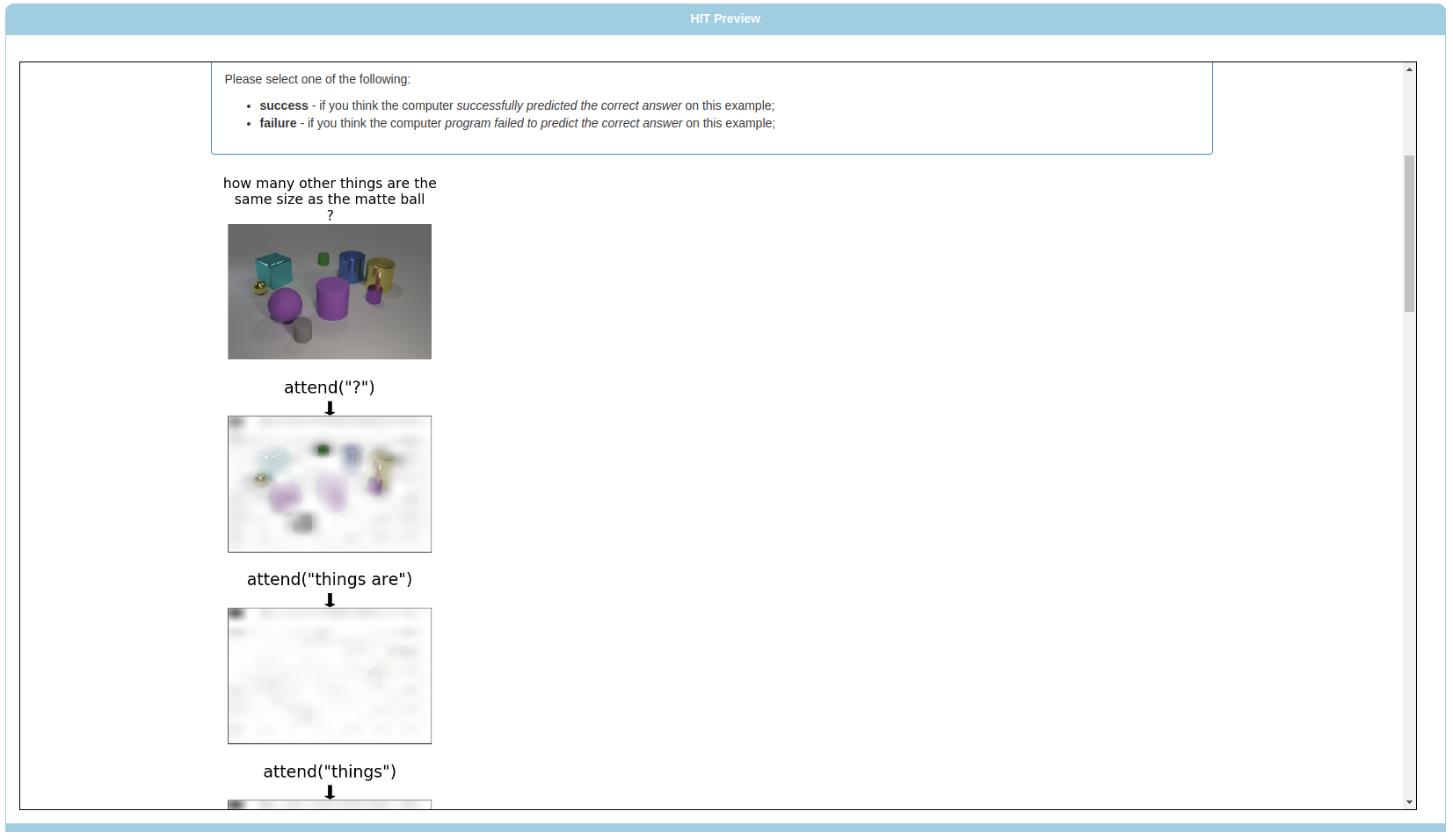} \\
\includegraphics[height=.3\textheight]{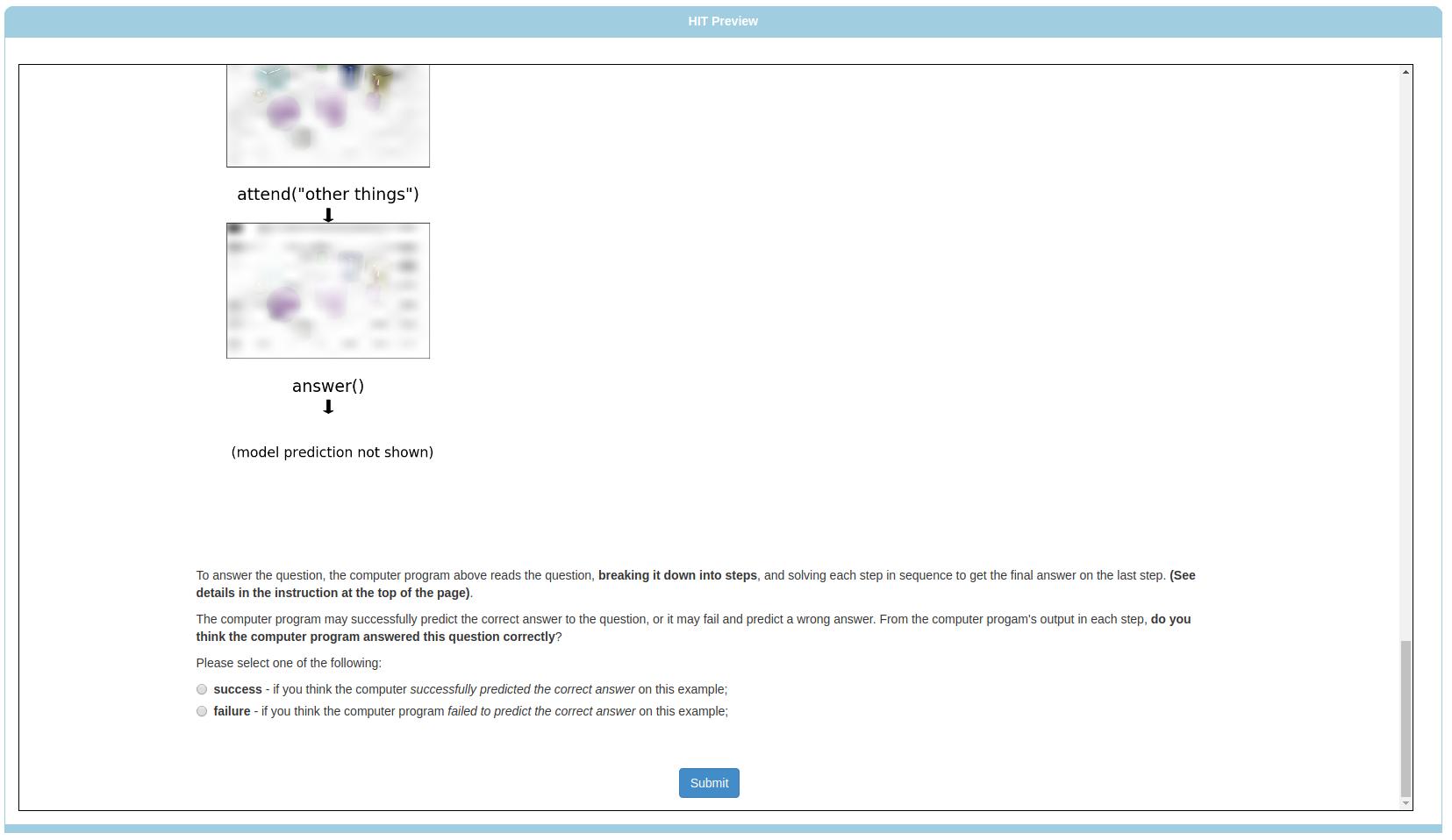}
\caption{The evaluation interface for \textit{forward prediction (failure detection)} on MAC \cite{hudson2018compositional} model, deployed on Amazon Mechanical Turk (AMT).}
\label{fig:amt_failure_detection_mac}
\end{figure}

In our human evaluation described in Sec. 4.2 of the main paper, we deployed the evaluation on the Amazon Mechanical Turk (AMT) platform. For subjective understanding, we showed 200 visualized examples of each model, and asked the human evaluators to judge whether the model's reasoning steps were clear and understandable to them. The evaluation interface is shown in Figure \ref{fig:amt_subjective_understanding_ours} and \ref{fig:amt_subjective_understanding_mac}. For forward prediction (failure detection), we collected 100 successful and 100 failure examples from each model, and the human evaluators were asked to predict whether the model succeeds or fails on each example. The evaluation interface is shown in Figure \ref{fig:amt_failure_detection_ours} and \ref{fig:amt_failure_detection_mac}. In both subjective understanding and forward prediction evaluation, we paid the AMT workers \$0.5 USD for each example they annotated.

\begin{figure}[t]
\center
\footnotesize
\begin{tabular}{c|c|cc}
ours with & ours without & \multicolumn{2}{c}{MAC \cite{hudson2018compositional} with 12 steps} \\
expert layout & expert layout & Step 1 to 6 & Step 7 to 12 \\
\hline
\includegraphics[trim={0 0 0 0},clip,scale=.18,valign=T]{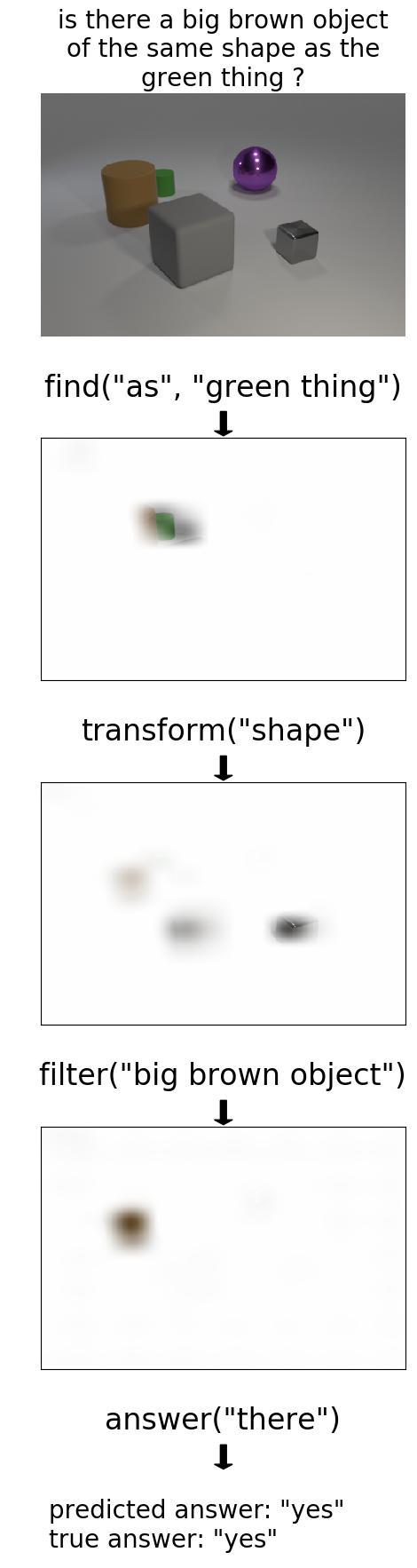} &
\includegraphics[trim={0 0 0 0},clip,scale=.18,valign=T]{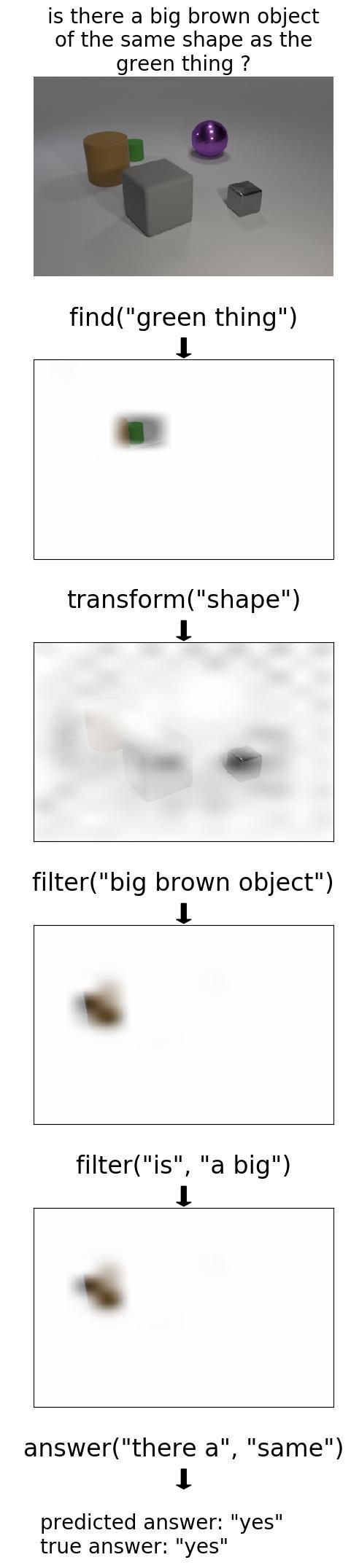} &
\includegraphics[trim={0 0 0 0},clip,scale=.18,valign=T]{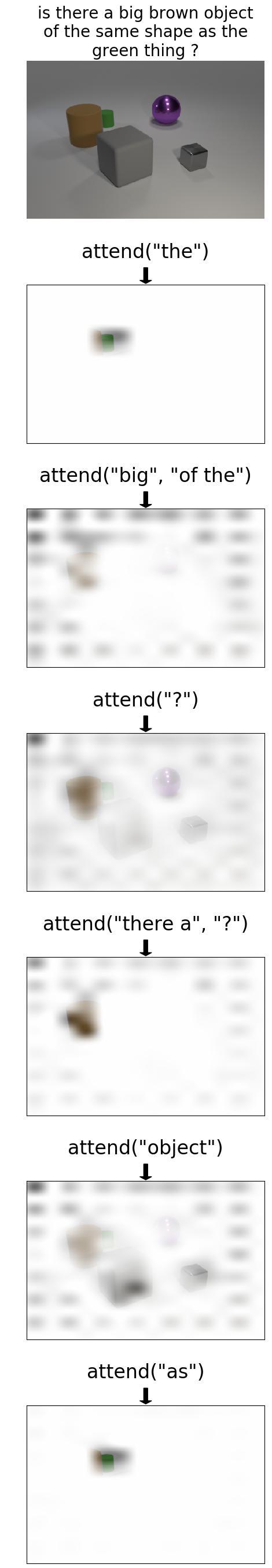} &
\includegraphics[trim={0 0 0 0},clip,scale=.18,valign=T]{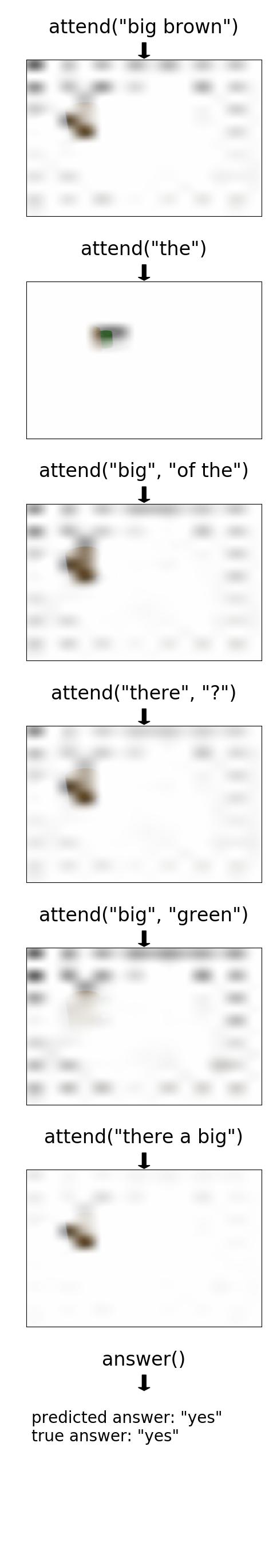} \\
\end{tabular}
\end{figure}

\begin{figure}[t]
\center
\footnotesize
\begin{tabular}{c|c|cc}
ours with & ours without & \multicolumn{2}{c}{MAC \cite{hudson2018compositional} with 12 steps} \\
expert layout & expert layout & Step 1 to 6 & Step 7 to 12 \\
\hline
\includegraphics[trim={0 0 0 0},clip,scale=.18,valign=T]{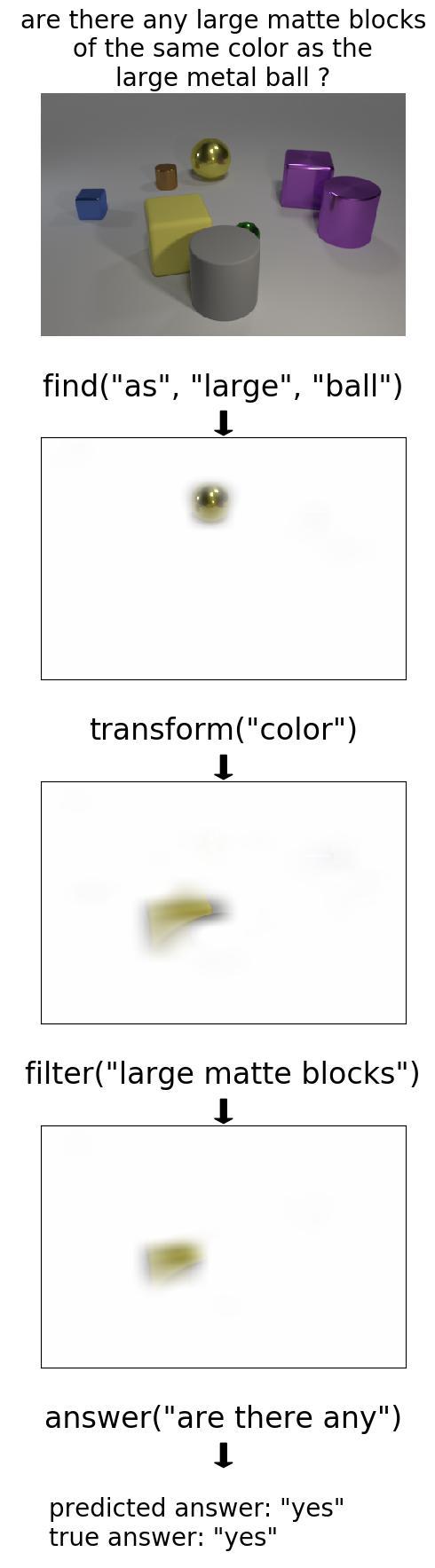} &
\includegraphics[trim={0 0 0 0},clip,scale=.18,valign=T]{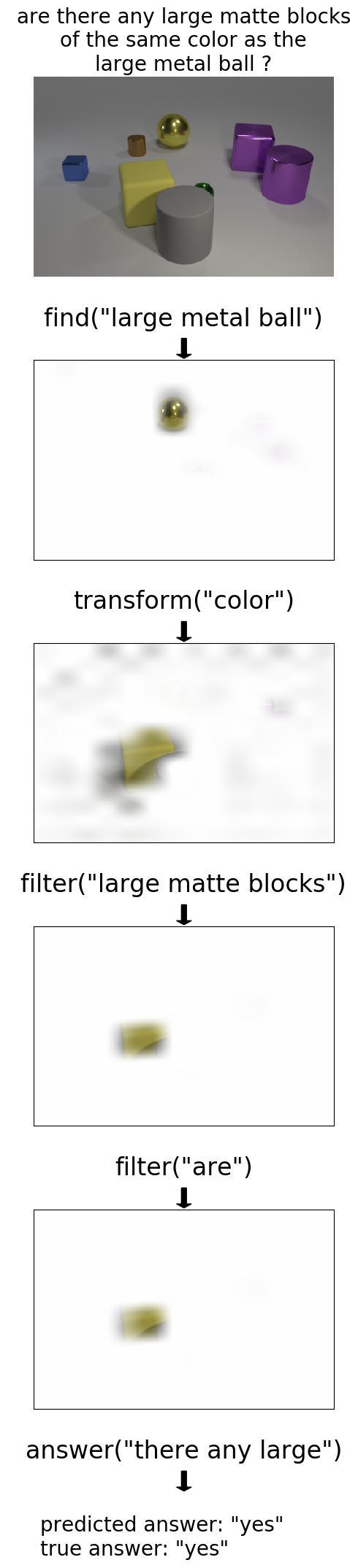} &
\includegraphics[trim={0 0 0 0},clip,scale=.18,valign=T]{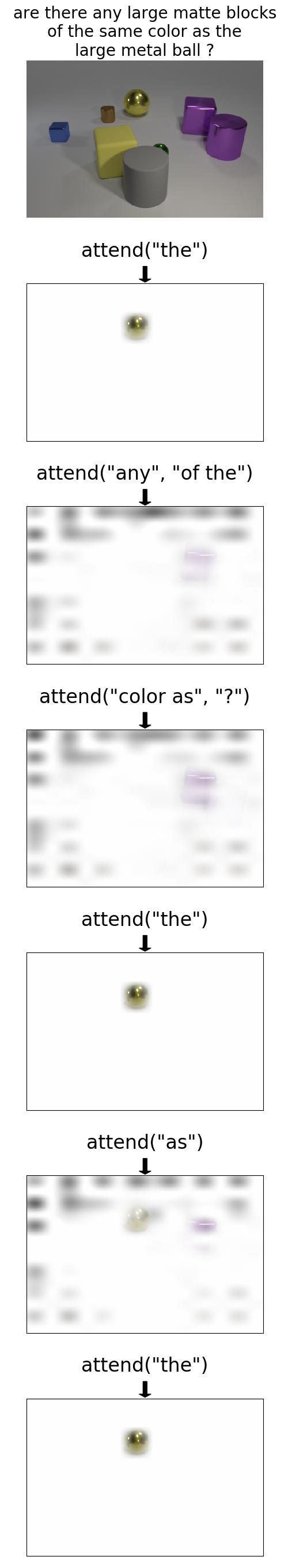} &
\includegraphics[trim={0 0 0 0},clip,scale=.18,valign=T]{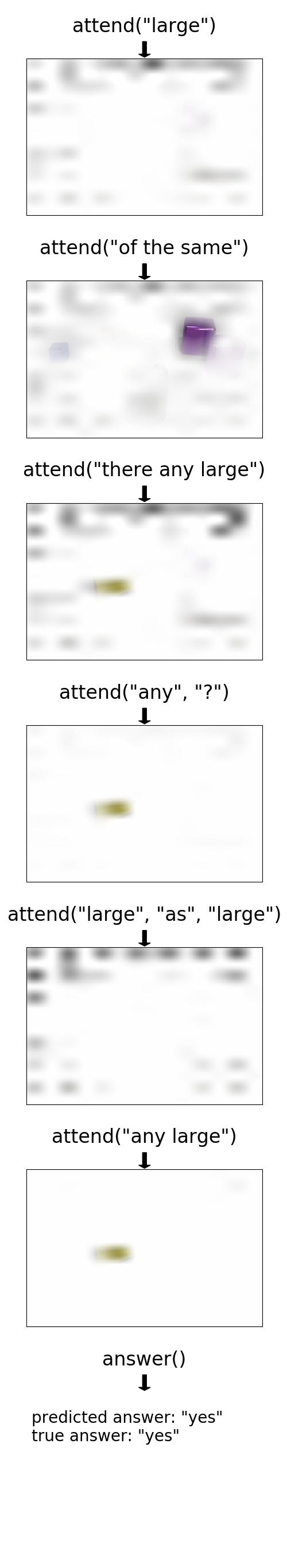} \\
\end{tabular}
\end{figure}

\begin{figure}[t]
\center
\footnotesize
\begin{tabular}{c|c|cc}
ours with & ours without & \multicolumn{2}{c}{MAC \cite{hudson2018compositional} with 12 steps} \\
expert layout & expert layout & Step 1 to 6 & Step 7 to 12 \\
\hline
\includegraphics[trim={0 0 0 0},clip,scale=.18,valign=T]{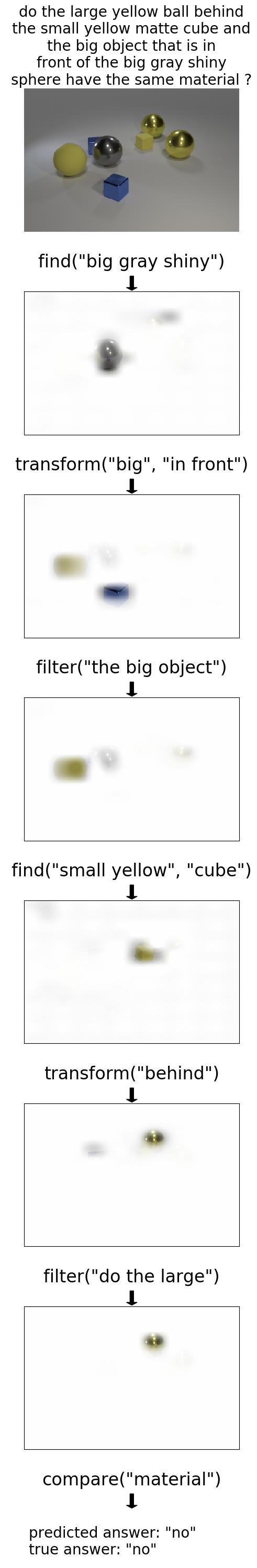} &
\includegraphics[trim={0 0 0 0},clip,scale=.18,valign=T]{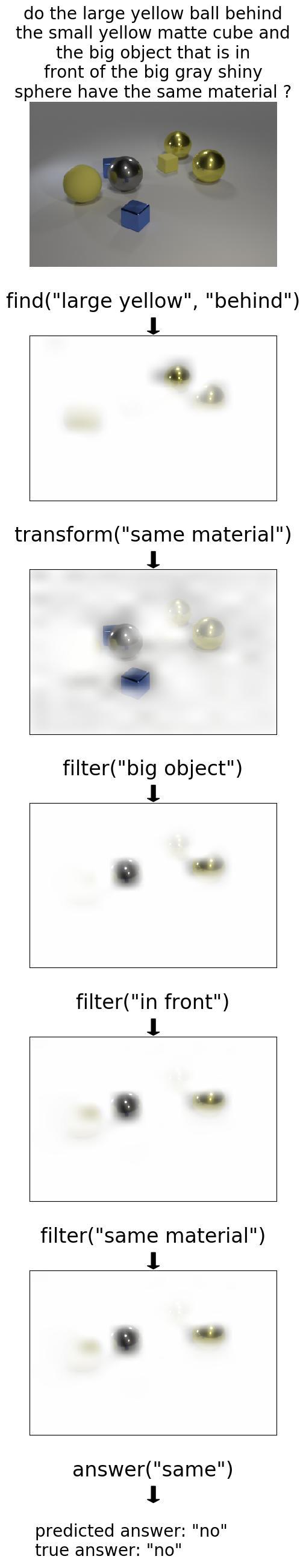} &
\includegraphics[trim={0 0 0 0},clip,scale=.18,valign=T]{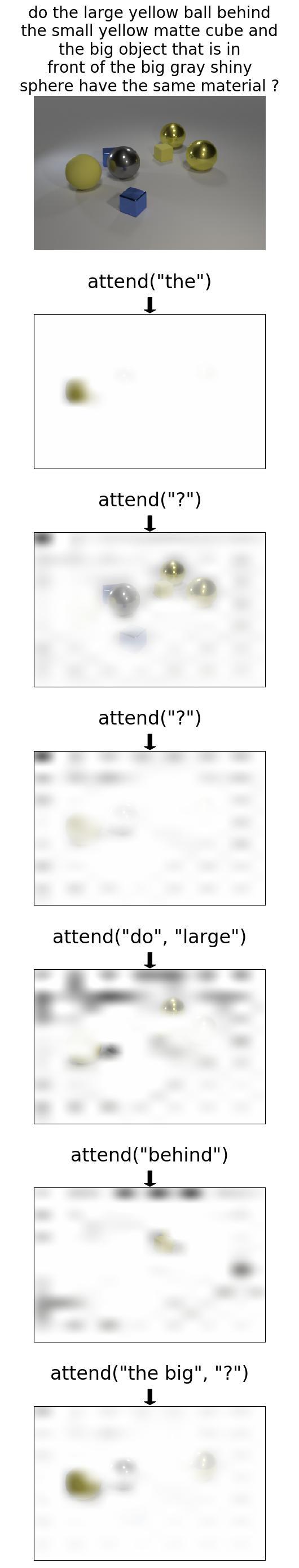} &
\includegraphics[trim={0 0 0 0},clip,scale=.18,valign=T]{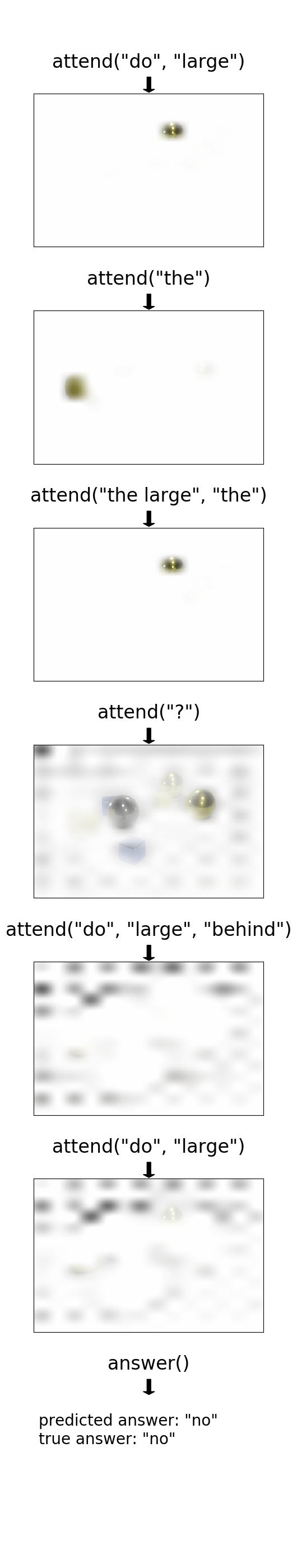} \\
\end{tabular}
\end{figure}

\begin{figure}[t]
\center
\footnotesize
\begin{tabular}{c|c|cc}
ours with & ours without & \multicolumn{2}{c}{MAC \cite{hudson2018compositional} with 12 steps} \\
expert layout & expert layout & Step 1 to 6 & Step 7 to 12 \\
\hline
\includegraphics[trim={0 0 0 0},clip,scale=.18,valign=T]{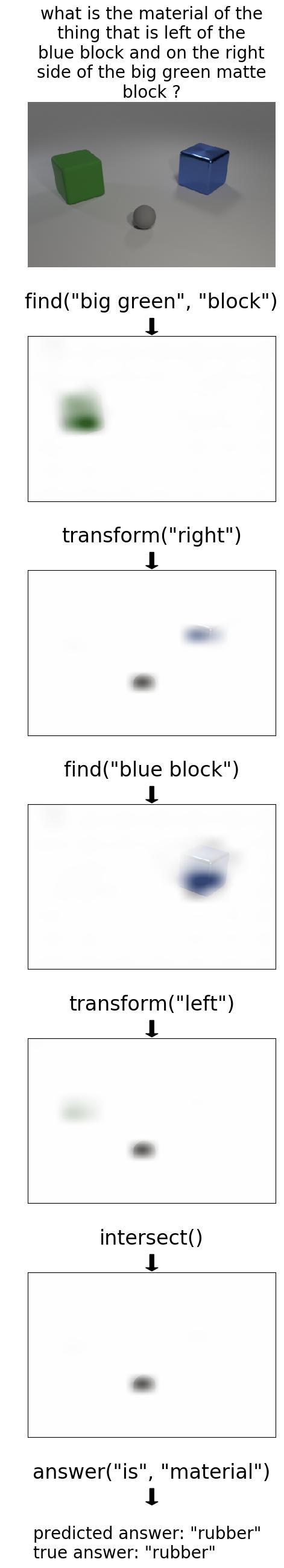} &
\includegraphics[trim={0 0 0 0},clip,scale=.18,valign=T]{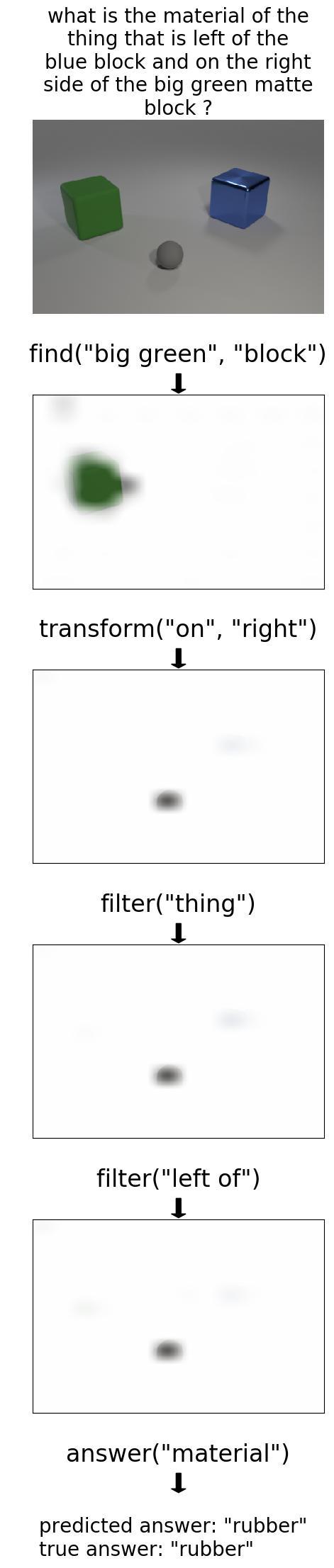} &
\includegraphics[trim={0 0 0 0},clip,scale=.18,valign=T]{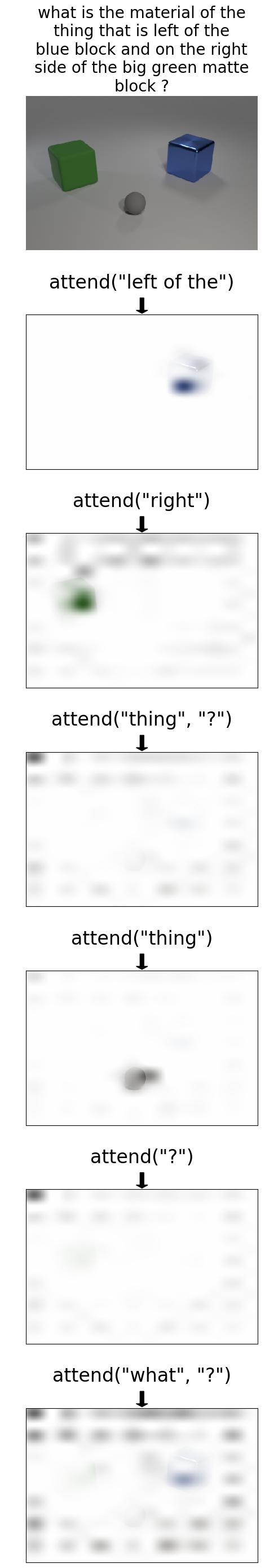} &
\includegraphics[trim={0 0 0 0},clip,scale=.18,valign=T]{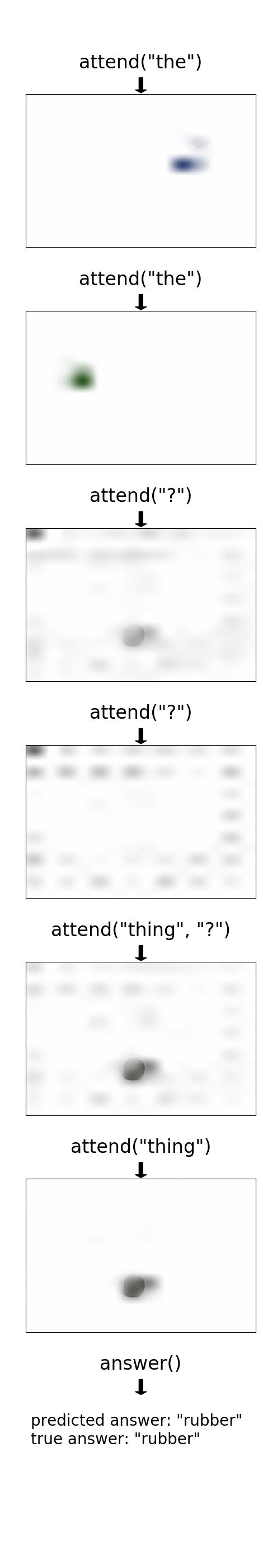} \\
\end{tabular}
\end{figure}

\begin{figure}[t]
\center
\footnotesize
\begin{tabular}{c|c|cc}
ours with & ours without & \multicolumn{2}{c}{MAC \cite{hudson2018compositional} with 12 steps} \\
expert layout & expert layout & Step 1 to 6 & Step 7 to 12 \\
\hline
\includegraphics[trim={0 0 0 0},clip,scale=.18,valign=T]{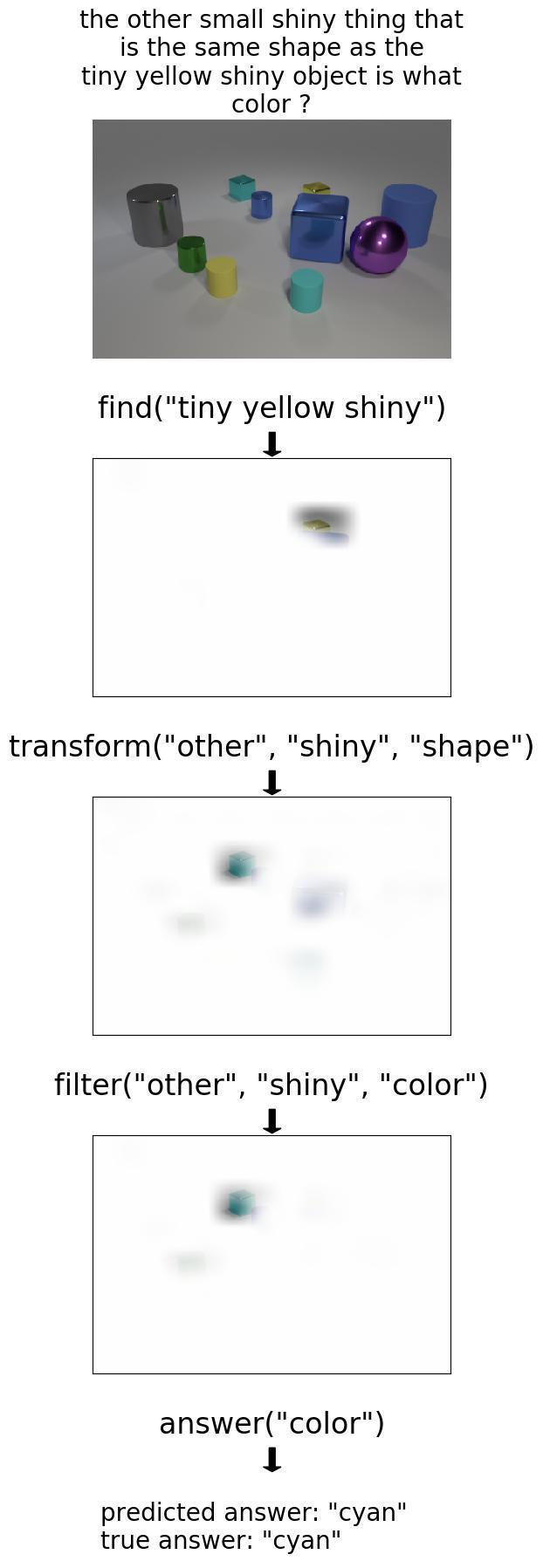} &
\includegraphics[trim={0 0 0 0},clip,scale=.18,valign=T]{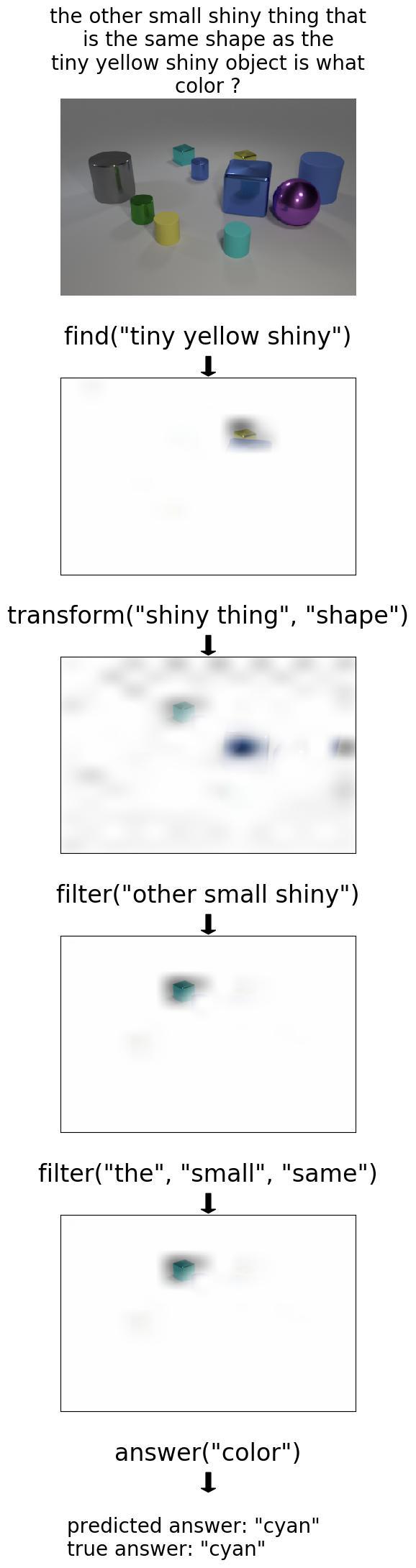} &
\includegraphics[trim={0 0 0 0},clip,scale=.18,valign=T]{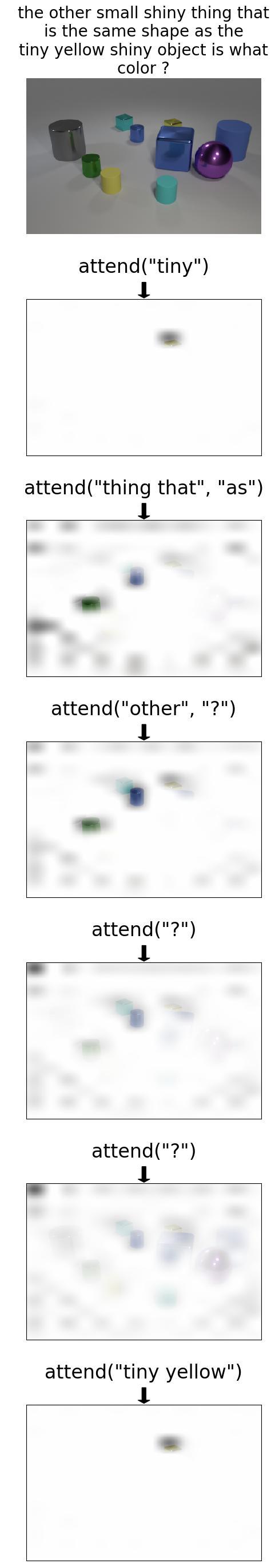} &
\includegraphics[trim={0 0 0 0},clip,scale=.18,valign=T]{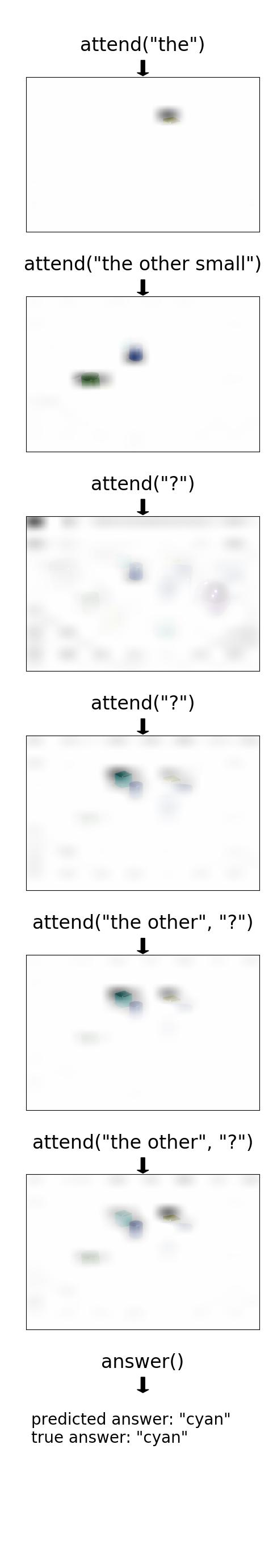} \\
\end{tabular}
\end{figure}

\begin{figure}[t]
\center
\footnotesize
\begin{tabular}{c|c|cc}
ours with & ours without & \multicolumn{2}{c}{MAC \cite{hudson2018compositional} with 12 steps} \\
expert layout & expert layout & Step 1 to 6 & Step 7 to 12 \\
\hline
\includegraphics[trim={0 0 0 0},clip,scale=.18,valign=T]{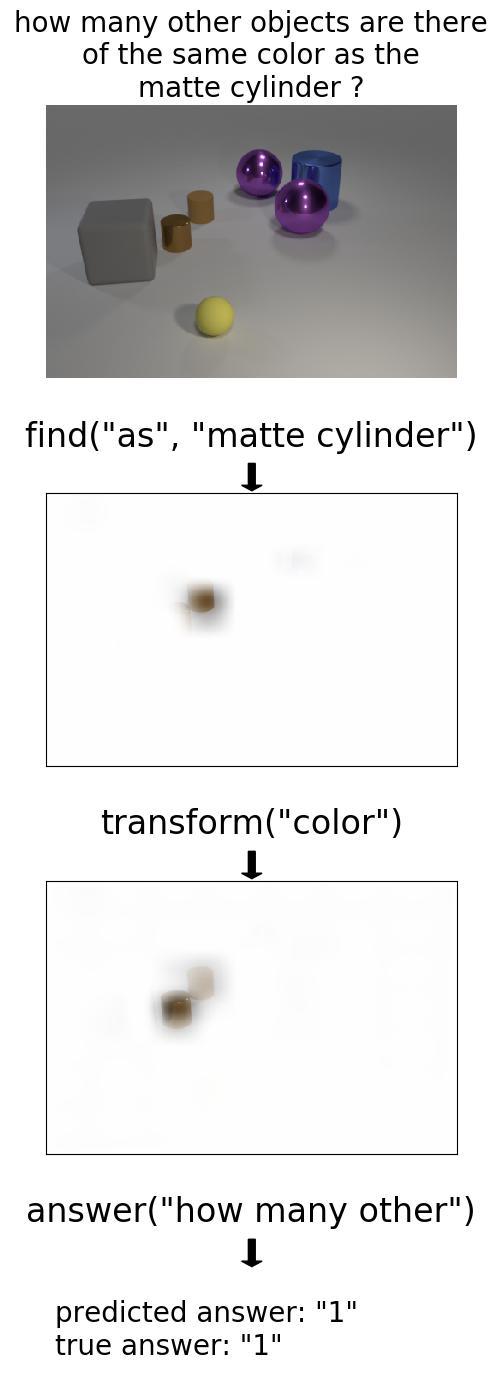} &
\includegraphics[trim={0 0 0 0},clip,scale=.18,valign=T]{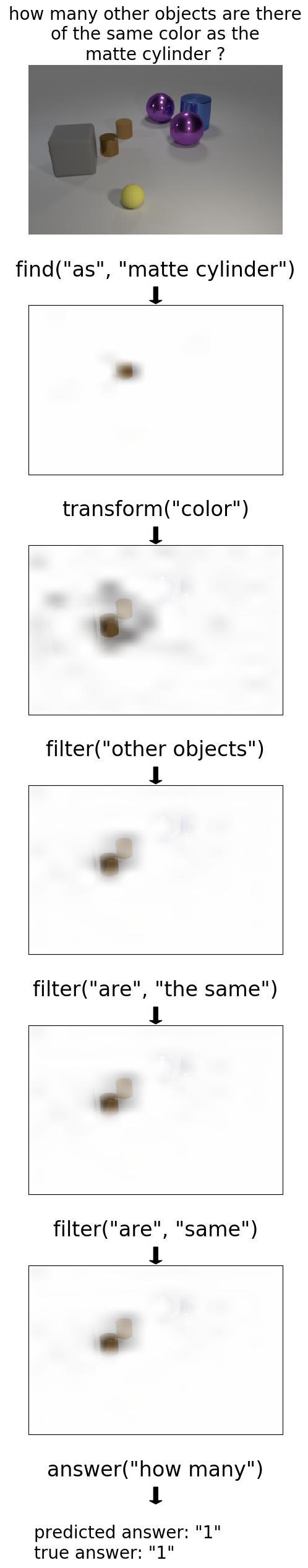} &
\includegraphics[trim={0 0 0 0},clip,scale=.18,valign=T]{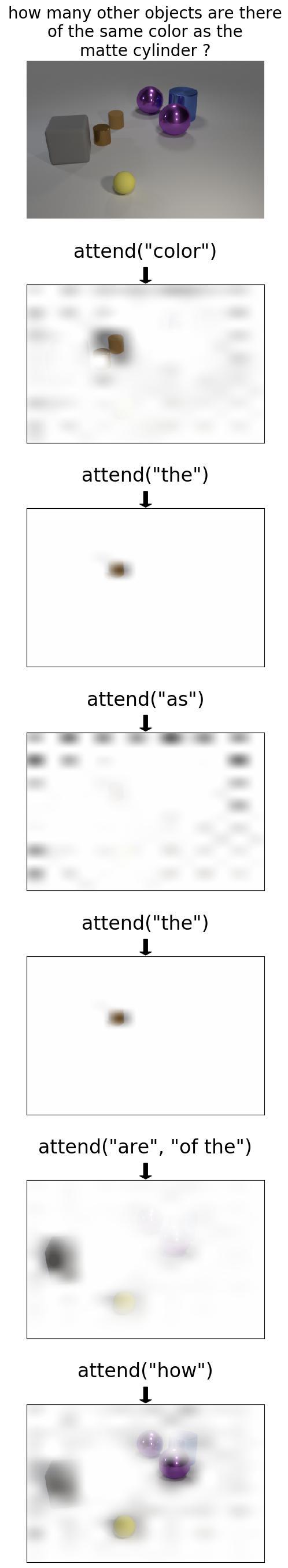} &
\includegraphics[trim={0 0 0 0},clip,scale=.18,valign=T]{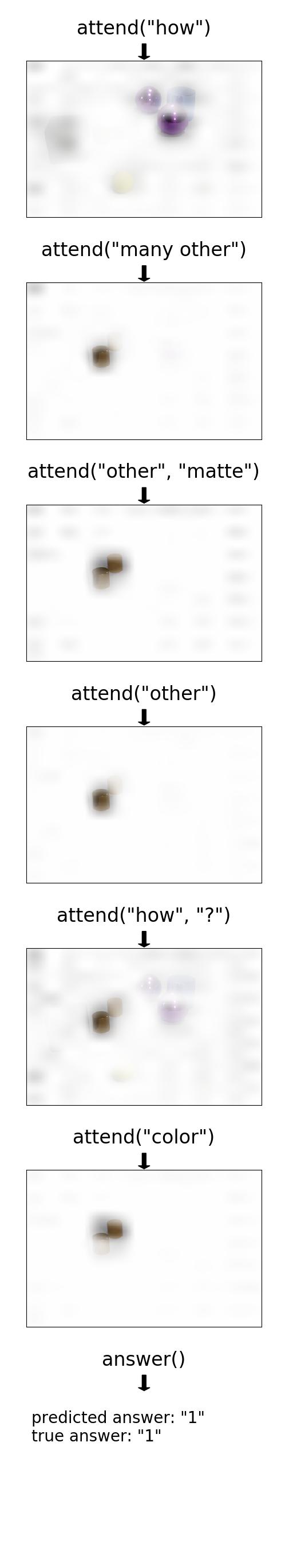} \\
\end{tabular}
\end{figure}

\end{document}